\documentclass{article}
\usepackage{graphicx} 
\usepackage[utf8]{inputenc}
\usepackage{authblk}
\usepackage[numbers]{natbib} 
\usepackage{xcolor}
\usepackage{amsmath} 
\usepackage{amsmath,amssymb,amsthm}
\usepackage{graphicx}
\usepackage{svg} 
\usepackage{fullpage}
\usepackage[skip=1em]{parskip}
\usepackage{enumitem}
\usepackage{placeins}
\usepackage{multirow}
\usepackage{float}
\usepackage{svg}
\usepackage{multicol}
\usepackage{amsmath,amsthm}
\usepackage{mdframed}
\usepackage{subcaption}
\usepackage{fontawesome5}
\usepackage[flushleft]{threeparttable} 
\usepackage{longtable}
\usepackage{booktabs}
\usepackage{tikz}
\usetikzlibrary{calc}
\usepackage{caption}    
\usepackage[margin=1in]{geometry}
\usepackage{url} 
\usepackage{longtable}
\usepackage{booktabs}
\usepackage{pdflscape}
\usepackage{array}
\usepackage{mathtools}
\usepackage{hyperref}
\usepackage{nicefrac}       
\usepackage{microtype}      
\usepackage{xcolor}         

\title{G-PARC: Graph-Physics Aware Recurrent Convolutional Neural Networks for Spatiotemporal Dynamics on Unstructured Meshes}
\author[1]{Jack T. Beerman}
\author[1]{Tyler J. Abele}
\author[2]{Mehdi Taghizadeh}
\author[3]{Andrew Davis}
\author[1]{Zo\"{e} J. Gray}
\author[2]{Negin Alemazkoor}
\author[3]{Xinfeng Gao}
\author[4]{H.S. Udaykumar}
\author[1,3,*]{Stephen S. Baek}
\affil[1]{\small School of Data Science, University of Virginia, Charlottesville, VA 22903, United States}
\affil[2]{\small Department of Engineering Systems and Environment, University of Virginia, Charlottesville, VA 22903, United States}
\affil[3]{\small Department of Mechanical and Aerospace Engineering, University of Virginia, Charlottesville, VA 22903, United States}
\affil[4]{\small Department of Mechanical Engineering, University of Iowa, Iowa City, IA 52242, United States}
\affil[*]{Corresponding author: baek@virginia.edu}

\begin{document}
\maketitle

\begin{center}
\href{https://github.com/JackBeerman/G-PARC}{\faGithub\ \textbf{Github}}
\end{center}
\begin{abstract}
Physics-aware recurrent convolutional networks (PARC) have demonstrated strong performance in predicting nonlinear spatiotemporal dynamics by embedding differential operators directly into the computational graph of a neural network. However, pixel-based convolutions are restricted to static, uniform Cartesian grids, making them ill-suited to following evolving localized structures in an efficient manner. Graph neural networks (GNNs) naturally handle irregular spatial discretizations, but existing graph-based physics-aware deep learning (PADL) methods have difficulty handling extreme nonlinear regimes. To address these limitations, we propose Graph PARC (G-PARC), which uses moving least squares (MLS) kernels to approximate spatial derivatives on unstructured graphs, and embeds the derivatives of governing partial differential equations into the network's computational graph. G-PARC achieves better accuracy with 2-3$\times$ fewer parameters than MeshGraphNet, MeshGraphKAN, and GraphSAGE, replacing the traditional encoder-processor-decoder framework with analytically computed differential operators. We demonstrate that G-PARC (1) generalizes across nonuniform spatial and temporal discretizations; (2) handles moving meshes required for structural deformation; and (3) outperforms existing graph-based PADL methods on nonlinear benchmarks including fluvial hydrology, planar shock waves, and elastoplastic dynamics.  By embedding explicit physical operators within the flexibility of GNNs, G-PARC enables accurate modeling of extreme nonlinear phenomena on complex computational domains, moving PADL beyond idealized Cartesian grids.

\end{abstract}

\section{Introduction}

Physics-aware recurrent convolutional neural networks (PARC) have demonstrated strong performance in nonlinear regimes of spatiotemporal dynamics, in which complex features such as shocks, extreme localized values, and coupled evolving fields are prevalent~\cite{nguyen2023parc, nguyen2024parcv2}. PARC has proven particularly effective at capturing discontinuities, fast transients, and sharp gradients in computational fluid dynamics (CFD), with sparse training data. This effectiveness stems from the neural architecture’s adherence to numerical principles: PARC embeds governing PDEs directly into its computational graph via explicit time integration and spatial differentiation operations, producing physically grounded predictions without solely relying on ``black-box” data-driven learning. Moreover, this approach has yielded a suite of models for highly nonlinear spatiotemporal dynamics~\cite{cheng2024physics, cheng2025multi, gray2025reduced} and has motivated the incorporation of classical numerical techniques into neural architecture design~\cite{beerman2026size}.

However, PARC's formulation remains fundamentally constrained to uniform Euclidean domains and constant temporal sampling. Consequently, PARC is not ideally suited  for modeling the non-uniform spatiotemporal dynamics characteristic of real-world applications---localized physics at shocks, interfaces, etc., irregular geometries, and variable timestep sizes.

Graph neural networks (GNNs) have become the preferred architecture for modeling physical systems on non-Euclidean domains. MeshGraphNets (MGNET) have demonstrated strong performance on complex, highly nonlinear systems involving contact and fluids~\cite{pfaff2020learning}. Unlike PARC, these models do not explicitly encode differential equations or physical laws; instead, they learn interaction dynamics entirely from data through deep message passing architectures with large amounts of data via an encoder-processor-decoder architecture. While this flexibility enables the simulation of diverse physical phenomena on unstructured meshes, it amplifies concerns regarding the black-box nature of AI/ML and lack of physical interpretability.

Several approaches have attempted to incorporate physical interpretability into graph-based architectures. Recent variations of MGNETs have leveraged Kolmogorov-Arnold Networks (KAN)~\cite{liu2024kan} with the global conservation laws in the network’s loss function and spline-based activation functions, yielding HydroGraphNet or MeshGraphKan (MGKAN)~\cite{taghizadeh2025interpretable}. Physics-informed graph neural networks (PiGNN) use moving least squares (MLS) to construct physics-informed loss functions ~\cite{zhang2025combining}. Fourier Neural Operators (FNO), originally limited to regular grids \cite{li2020fourier}, have been extended to geometry-informed neural operators (GINO) and graph neural operators (GNO) by combining graph-based mappings with operator learning in the latent space \cite{li2023geometry}. However, these extensions inherit the fundamental limitations of their parent physics-informed neural networks (PINN) and neural operators (NO) formulations. PINNs suffer from convergence issues and struggle to generalize outside the training domain~\cite{NEURIPS2021_df438e52, wang2021understanding} while NOs face aliasing errors that corrupt high-frequency details, spectral bias which favors capturing low-frequency features, and high computational costs \cite{fanaskov2023spectral, bartolucci2023neural}.

\begin{figure}
    \centering
    \includegraphics[width=\linewidth]{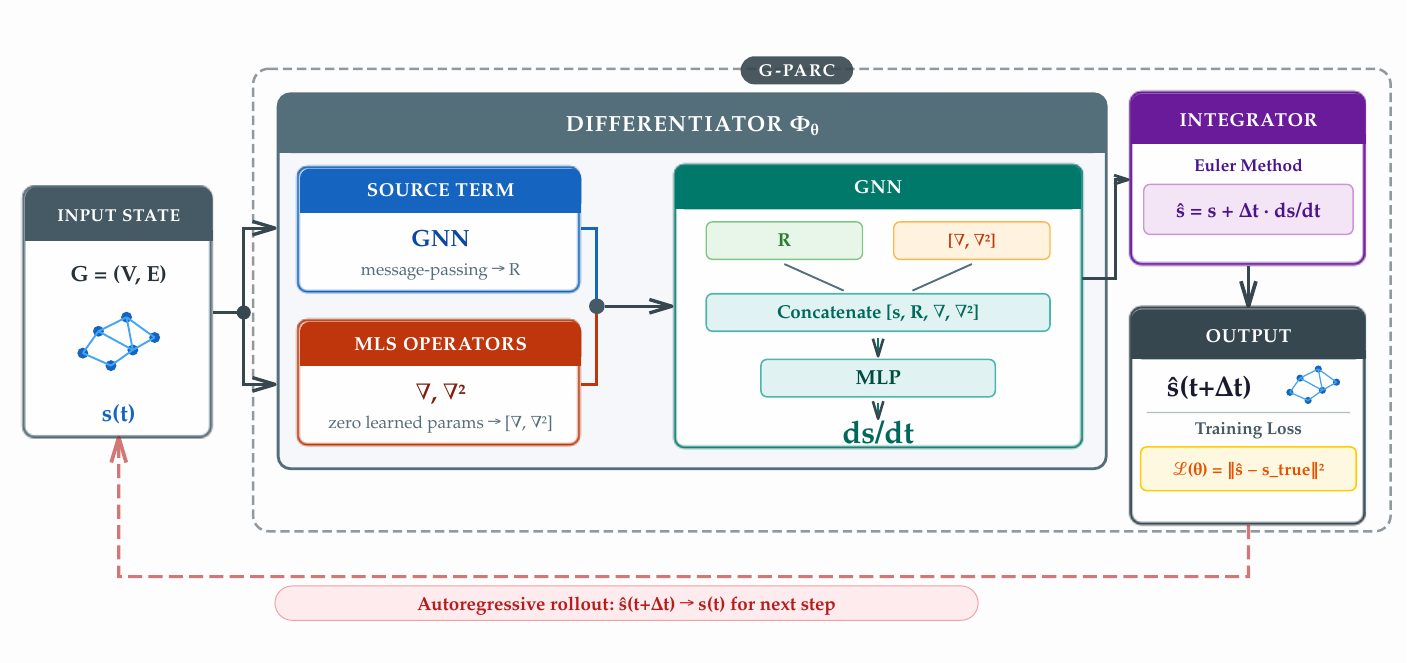}
    \caption{\textbf{G-PARC architecture with MLS differential operators.} A GNN approximates source terms $\boldsymbol{R}$, while domain-specific MLS operators compute physics features $\nabla, \nabla^2$ with zero learned parameters. The fusion module concatenates $[s, \nabla, \nabla^2, \boldsymbol{R}]$ and approximates $d\boldsymbol{s}/dt$ via an MLP, which is then advanced by a numerical integrator (Euler/Heun/RK4) and fed back autoregressively.}
    \label{fig:gparc_architecture}
\end{figure}

Building on PARC's strong performance relative to PINNs and NOs demonstrated in ~\cite{nguyen2023parc, nguyen2024parcv2, cheng2024physics}, we introduce Graph PARC (G-PARC) to extend PARC's capabilities to irregular geometries. G-PARC employs physics-aware graph message passing to handle spatially irregular discretizations while incorporating variable timestep sizes and global physical parameters. Critically, G-PARC constructs differential operators \textbf{directly within} the message-passing framework using moving least squares (MLS), computing geometry-derived stencil coefficients at each node analogous to least-squares gradient reconstruction in unstructured finite volume methods, ensuring that physical structure is embedded in the representation before any network weights are applied (\textbf{Figures~\ref{fig:gparc_architecture} and~\ref{fig:mlsdiagram}}). Moreover, the analytical representation of differential operators within the computational graph of the network dramatically increases computational efficiency by removing the traditional encoder-processor-decoder framework in graph PADL approaches.

This work presents the first extension of PARC to non-Euclidean domains and demonstrates the efficacy of G-PARC across three distinct spatiotemporal regimes. First, we evaluate fluvial hydrology simulations over irregular, LiDAR-captured terrain to establish spatial generalization on real-world geometries. Second, we consider compressible shock tube flows to validate temporal generalization, demonstrating that conditioning on timestep size ($\Delta t$) and global parameters enable robust prediction across varying sampling rate and physical regimes. Finally, we apply G-PARC to elastoplastic structural mechanics on deforming Lagrangian meshes---overcoming the fixed-stencil limitations of CNNs. Our results show that G-PARC generalizes across nonuniform, highly nonlinear spatiotemporal dynamics while maintaining the interpretability, efficiency, and fidelity required to advance PARC as a robust, trustworthy neural surrogate solver for scientific discovery.

\section{Results}

We evaluate G-PARC across three benchmarks, each chosen to isolate a distinct generalization challenge that existing PADL methods have not simultaneously addressed. First, fluvial hydrology simulations over irregular, LiDAR-captured terrain test whether G-PARC’s MLS operators can  accurately approximate differential quantities on unstructured meshes, where non-uniform cells and complex topography make local operator approximation significantly harder than a regular grid. Second, compressible shock tube flows on a uniform Euclidean domain, isolates temporal generalization with cases spanning a $3.7\times$ range of simulation timesteps and varying left-state conditions that require the model to adapt its predicted dynamics to simulation-specific CFL conditions while resolving sharp discontinuities in density, momentum, and energy. Third, the elastoplastic structural mechanics task from the PLAID benchmark~\cite{casenave2025physics} combines large-scale mesh structures (25--32k nodes), unique geometries, and prediction of nodal displacements during extreme deformation, including material damage and erosion where the effective geometry evolves as the structure deforms. 

Across all three domains, we benchmark against four models: G-PARC (\textbf{w/o MLS}) (the original formulation without MLS operators), MGKAN, and GraphSAGE~\cite{hamilton2017inductive}.  For the shock tube domain, we additionally compare PARCv2~\cite{nguyen2024parcv2}, the CNN-based predecessor to G-PARC, which requires regular grid structure and therefore cannot be applied to the irregular meshes of the fluvial hydrology and elastoplastic domains. We also attempted to benchmark against GNO and GINO; however, both were prohibitively expensive to train on an A100-80GB GPU (Table~\ref{tab:compute}). MGNET, MGKAN, GraphSAGE, GNO, and GINO all employ the encoder-processor-decoder framework introduced by Pfaff et al.~\cite{pfaff2020learning} and G-PARC eliminates this requirement when computing differential operators. Together, these baselines span the dominant approaches for learned mesh-based simulation, providing a comprehensive comparison across architectures. All approaches utilize a traditional data-driven loss.

To test the core contribution of G-PARC, we begin with irregular meshes from fluvial hydrology simulations obtained from the Hydrologic Engineering Center River Analysis Software (HEC-RAS)~\cite{HECRAS2025} integrated with United States Geological Survey (USGS) river flow data \cite{USGS2023NWIS}. The governing PDEs are the Shallow Water Equations (SWE), enforcing conservation of mass and momentum through advection of depth-average flow quantities and diffusion from turbulent mixing (Appendix~\ref{app:fluvial}). The dataset is derived from the Finite Volume Method (FVM), where each cell characterizes the underlying LiDAR-captured topography and captures dynamic field variables including water depth $\mathbf{h}$, volume $\mathbf{V}$, and velocity in the $x$ and $y$ directions $\mathbf{v}_x$, $\mathbf{v}_y$. G-PARC’s MLS operators analytically approximate these advection and diffusion terms directly on the irregular mesh neighborhoods. This dataset therefore provides a direct test of whether MLS-computed differential operators offer an advantage over learned message passing on unstructured geometries where cell sizes, and shapes vary significantly. 

\begin{figure}
    \centering
    \includegraphics[width=0.8\linewidth]{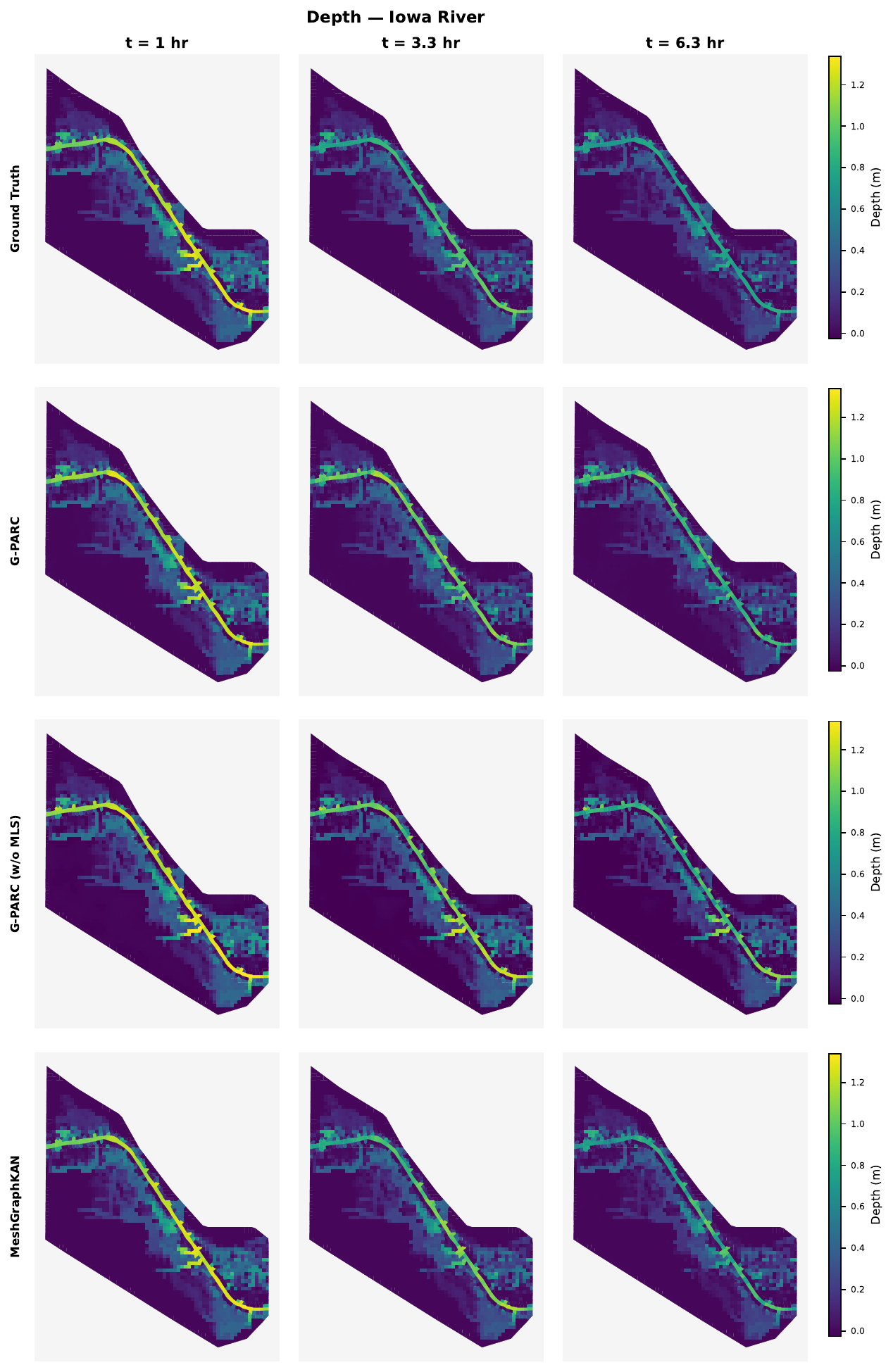}
    \caption{\textbf{Top three best performing model for Iowa River Flooding Model Prediction.}The models are predicting the water depth of the Iowa River utilizing a rollout prediction during a flood at 3 different time steps (t=1 hr, 3.3hrs, and 6.3 hrs). The top row is the ground truth, followed by G-PARC, G-PARC (\textbf{w/o MLS}), and MGKAN. All three models have a good  prediction at the t = 1hr mark, however, all models predict more flooding near the bottom of the river at time step t=3.3 hrs. G-PARC (\textbf{w/o MLS}) and MGKAN continue to have this issue at t=6.3 hrs, while G-PARC accurate predicts the stark decrease in water level throughout the River.
}
\label{fig:iowatop3}
\end{figure}

Figure~\ref{fig:iowatop3} depicts the top three models on the Iowa River and Table~\ref{tab:river} reports results computed across both the Iowa River and White River terrains with timesteps varying from 21 to 112 per simulation for all models. The vast variation in rollout length makes this benchmark particularly demanding for long-horizon rollout. G-PARC achieves the lowest RRMSE AUC ($0.208 \pm 0.106$) and NMSE AUC ($0.070 \pm 0.070$), a 31\% reduction in NMSE over the next-best model. SSIM and $R^2$ reach 0.952 and 0.909 respectively, indicating that G-PARC preserves both spatial structure and field variance throughout the rollout.

MGKAN is the best encoder-processor-decoder baseline, achieving the lowest final-timestep RRMSE (0.319) and competitive SSIM (0.947), though its higher NMSE (0.012) and lower $R^2$ (0.886) suggest that while individual frames appear visually plausible, cumulative error is greater than G-PARC. MGKAN’s strong performance among the baselines can be attributed to its Fourier KAN encoder which replaces standard linear weight parameters with learnable univariate functions, increasing the network’s representational capacity for nonlinear dynamics. Despite this rich feature representation, MGKAN still falls short of G-PARC, which benefits from explicit physics-aware operators that directly encode the governing spatial structure rather than relying on the network to learn it implicitly. 

Table~\ref{tab:river_pv} displays these metrics across the individual physical fields $\mathbf{h}$, $\mathbf{V}$, $\mathbf{v}_x$, $\mathbf{v}_y$. Additionally, the SWE conservation metrics established by Taghizadeh et al.\ \cite{taghizadeh2025interpretable}---capturing Critical Success Index (CSI) and Nash-Sutcliffe Efficiency (NSE)---are reported in Appendix Table~\ref{tab:fluvial_comparison} and reinforce the findings of the aggregate metrics presented here.

Having evaluated G-PARC on the SWEs with strong friction source terms and demonstrated its ability on irregular geometries, we now shift to a hyperbolic regime governed by the compressible Euler equations. Here, the MLS operators must capture conservation of fluxes across sharp discontinuities in density $\boldsymbol{\rho}$, pressure $\mathbf{p}$, and energy $\mathbf{E}$. Because this domain uses a uniform Euclidean grid, it also enables direct comparison to PARCv2. We consider a quasi-1D planar shock wave with varying global parameters and simulation-dependent timestep sizes, allowing us to assess G-PARC's adaptive temporal encoding and global parameter conditioning within a highly transient flow regime.

The shock tube is a quasi-1D problem discretized on a $\mathbf{64} \times \mathbf{64}$ grid over a \textbf{$0.5~\text{m} \times 0.5~\text{m}$} domain. The governing equations are the compressible Euler equations. The left state (L) corresponds to the high-pressure, high-density region, with a left-to-right pressure ratio of 10:1 and a left-to-right density ratio of 8:1. Global parameters---specifically the left-state $\boldsymbol{\rho}$, $\mathbf{p}$, and the simulation timestep $\Delta t$---are provided as conditioning inputs to the models.

\begin{figure}
    \centering
    \includegraphics[width=\linewidth]{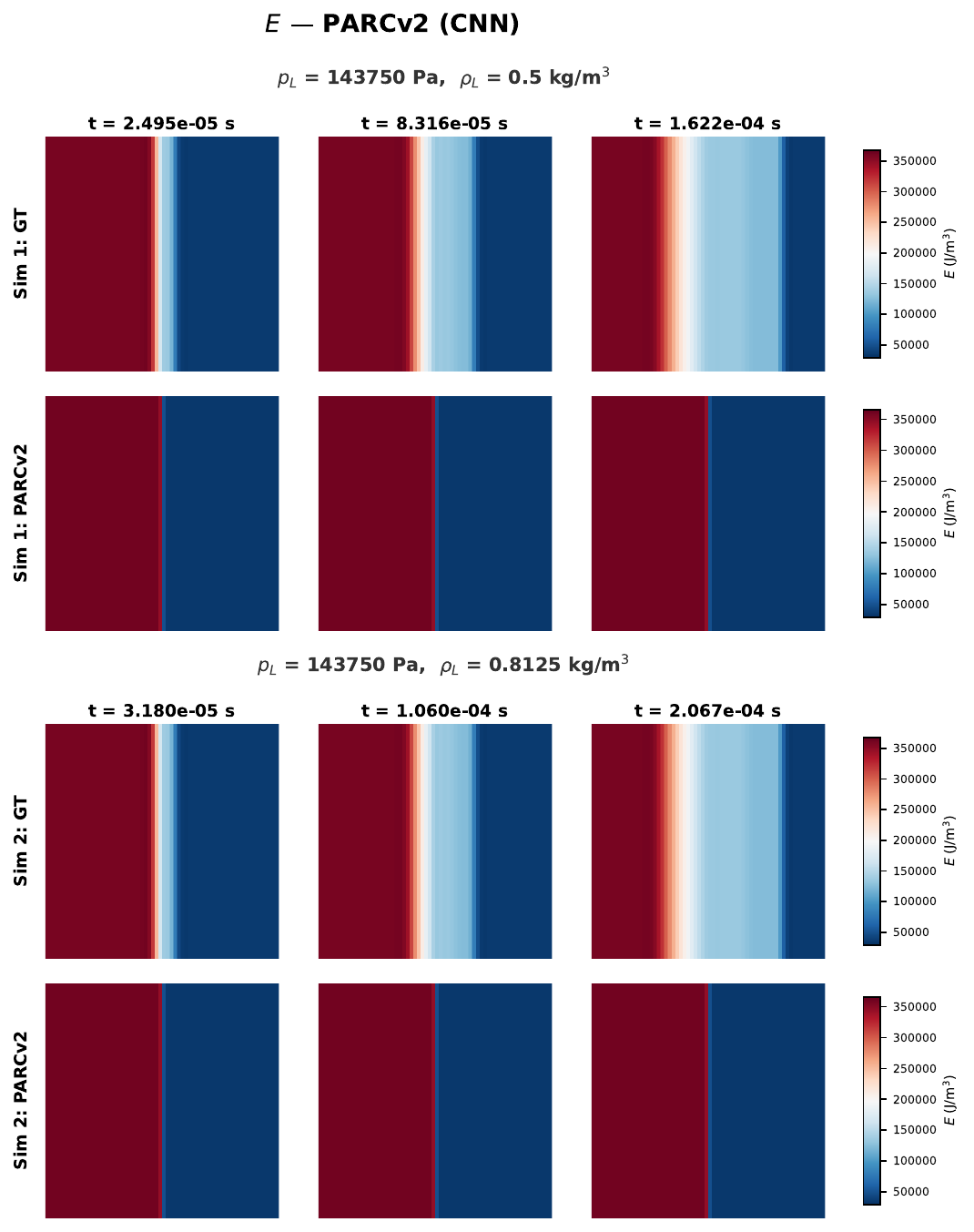}
   \caption{\textbf{Planar Shockwave Model Prediction for PARCv2.} PARCv2 predicts the total energy $E$ (J/m$^3$), which represents the sum of internal and kinetic energy per unit volume. PARCv2 fails to produce a meaningful prediction.}
    \label{fig:shockworst}
\end{figure}

Each simulation case is advanced with a constant timestep $\Delta t$ determined by the CFL condition, which depends on both $\mathbf{p}_\mathbf{L}$ and $\boldsymbol{\rho}_\mathbf{L}$ through the local sound speed. Across the 500 cases, $\Delta t$ varies by a factor of $3.7\times$, spanning from $3.84 \times 10^{-6}$ s to $1.41 \times 10^{-5}$ s. The full details of the train/validation/test split are provided in Appendix~\ref{app:shockdata}. A key limitation of PARCv2 is its inability to accommodate variable $\Delta t$ between simulations; temporal progression is implicitly assumed to be uniform across all training samples (Figure~\ref{fig:shockworst}). G-PARC addresses this by explicitly encoding $\Delta t$ as a global conditioning parameter alongside $\mathbf{p}_\mathbf{L}$ and $\boldsymbol{\rho}_\mathbf{L}$, enabling the model to adapt its predicted dynamics to simulation-specific temporal resolution.

\begin{figure}
    \centering
    \includegraphics[width=\linewidth]{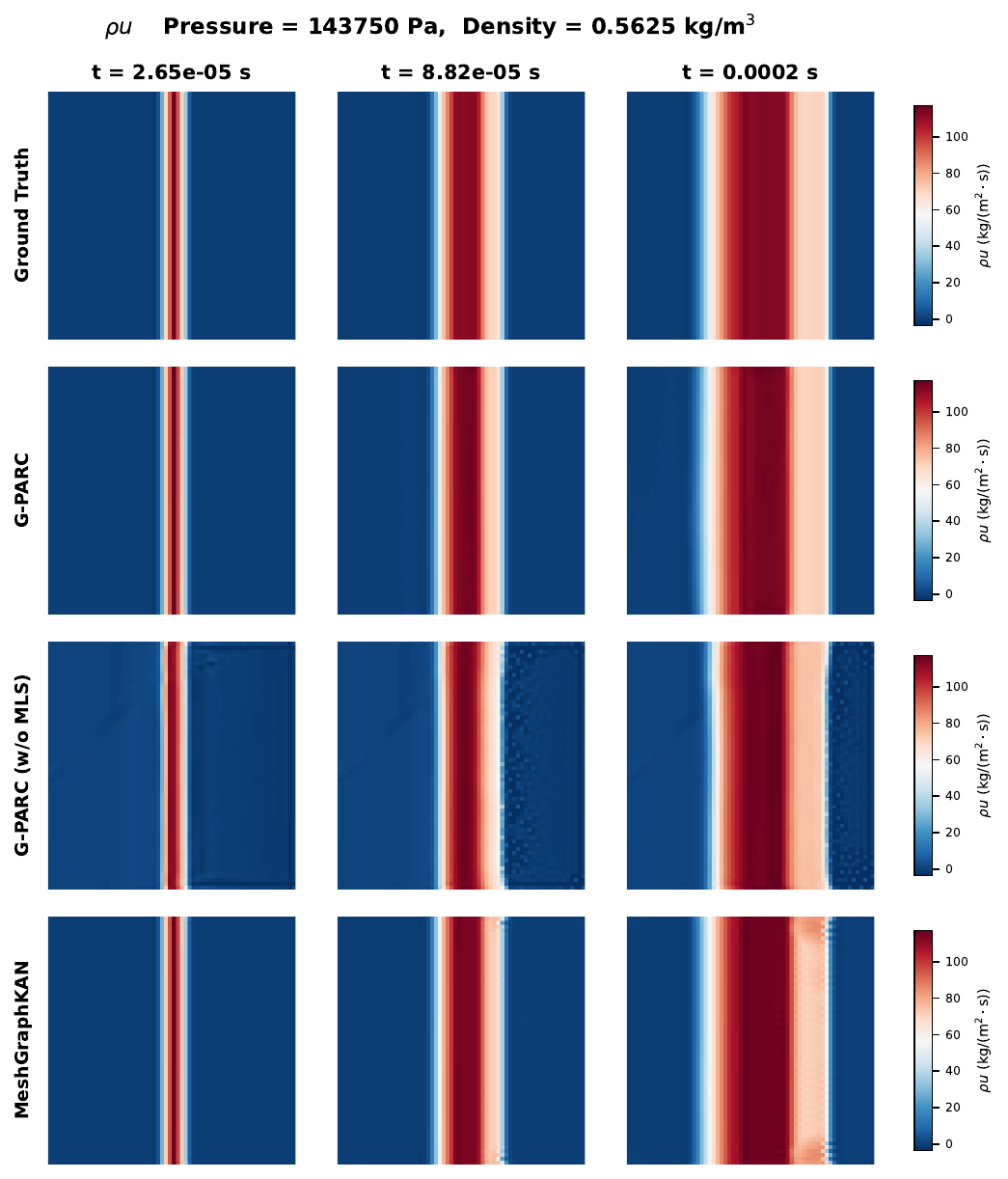}
    \caption{\textbf{Top three best performing models for Planar Shockwave Model Prediction.} The models are predicting x-momentum $\rho u$ with $\rho$ being fluid density (kg/m$^3$) and $u$ being velocity (m/s), meaning we measure $\rho u$ in terms of (kg/m$^2$s). We utilize a rollout prediction for all models at three different timesteps ($t = 2.65 \times 10^{-5}$ s, $t = 8.82 \times 10^{-5}$ s, and $t = 2.00 \times 10^{-4}$ s). Additionally, the pressure of the system is set to $143{,}750$ Pa and the density is set to $0.5625$ kg/m$^3$; these values remain constant across all models. The top row is the ground truth, followed by G-PARC, G-PARC (\textbf{w/o MLS}), and MGKAN. G-PARC demonstrates strong performance with great reconstruction while G-PARC (\textbf{w/o MLS}) and MGKAN exhibit multiple artifacts across time steps.
}
    \label{fig:shocktop}
\end{figure}

Table~\ref{tab:shock} summarizes model performance across six complementary metrics, while Figure~\ref{fig:shocktop} illustrates representative predictions of $x$-momentum for a selected test case ($\boldsymbol{p}_\mathbf{L} = 143{,}750$ Pa, $\mathbf{\rho}_\mathbf{L} = 0.5625$ kg/m$^3$). G-PARC achieves the best score on every metric, attaining an RRMSE AUC of 0.0070---a $7.9\times$ reduction compared to the next-best model, MGKAN (0.0555)---and an RRMSE\textsubscript{fin} of 0.0121, representing a $10.1\times$ improvement over G-PARC (\textbf{w/o MLS}) (0.1222). The NMSE AUC of $1.00\times10^{-4}$ is $265\times$ lower than MGKAN's 0.0393, while SSIM AUC reaches 0.9999 and $R^2$ reaches 0.9998, both near-perfect and substantially above G-PARC (\textbf{w/o MLS}) (0.9774 and 0.9704, respectively). Physical field results (Appendix Table~\ref{tab:shock_pv}) confirm that these improvements are consistent across $\boldsymbol{\rho}$, $\mathbf{p}$, and $\mathbf{E}$, with the largest gains observed in momentum prediction, which is the most challenging quantity due to steep gradients at the shock and contact discontinuities. MGNET, which is known to struggle with long-term rollout prediction, performs notably poorly for this flow than the other methods tested.

The gap between G-PARC (\textbf{w/o MLS}) and G-PARC further underscores the importance of embedding differential operators within the computational graph of the network. G-PARC (\textbf{w/o MLS}) achieves an overall RRMSE AUC of 0.0824, which, while substantially better than MGNET, still reflects difficulty in resolving the shock structure, particularly for the momentum field. The sole architectural distinction between G-PARC (\textbf{w/o MLS}) and G-PARC is the ability to explicitly approximate the divergence of flux terms of the compressible Euler equations. This provides the network with explicit representations of the flux structure that would otherwise need to be discovered implicitly through message passing alone.

Finally, we test G-PARC on the most complex non-linear dynamics of the three benchmarks: 2D elastoplastic structural mechanics under large deformations. This problem tests G-PARC’s ability to predict nodal displacement on deforming meshes and, critically, to generalize across simulations with unseen mesh topologies and node counts. We adopt the D-ElPlDynamics dataset from the PLAID benchmark suite, which simulates the transient response of 2D structures undergoing large elastoplastic deformations in a plane strain setting~\cite{casenave2025physics}. The dynamic nonlinear structural mechanics simulations are solved with OpenRadioss~\cite{openradioss2022} using the finite element method (FEM), which accounts for material damage through element erosion and failure. In each simulation, displacement is prescribed on the right boundary while the left boundary is held fixed. Crucially, the dataset varies the mesh topology across each simulation, making this benchmark a direct test of mesh-invariant generalization. The predicted quantities are the nodal displacement components $\mathbf{U}_x$ and $\mathbf{U}_y$.

\begin{figure}
    \centering
    \includegraphics[width=\linewidth]{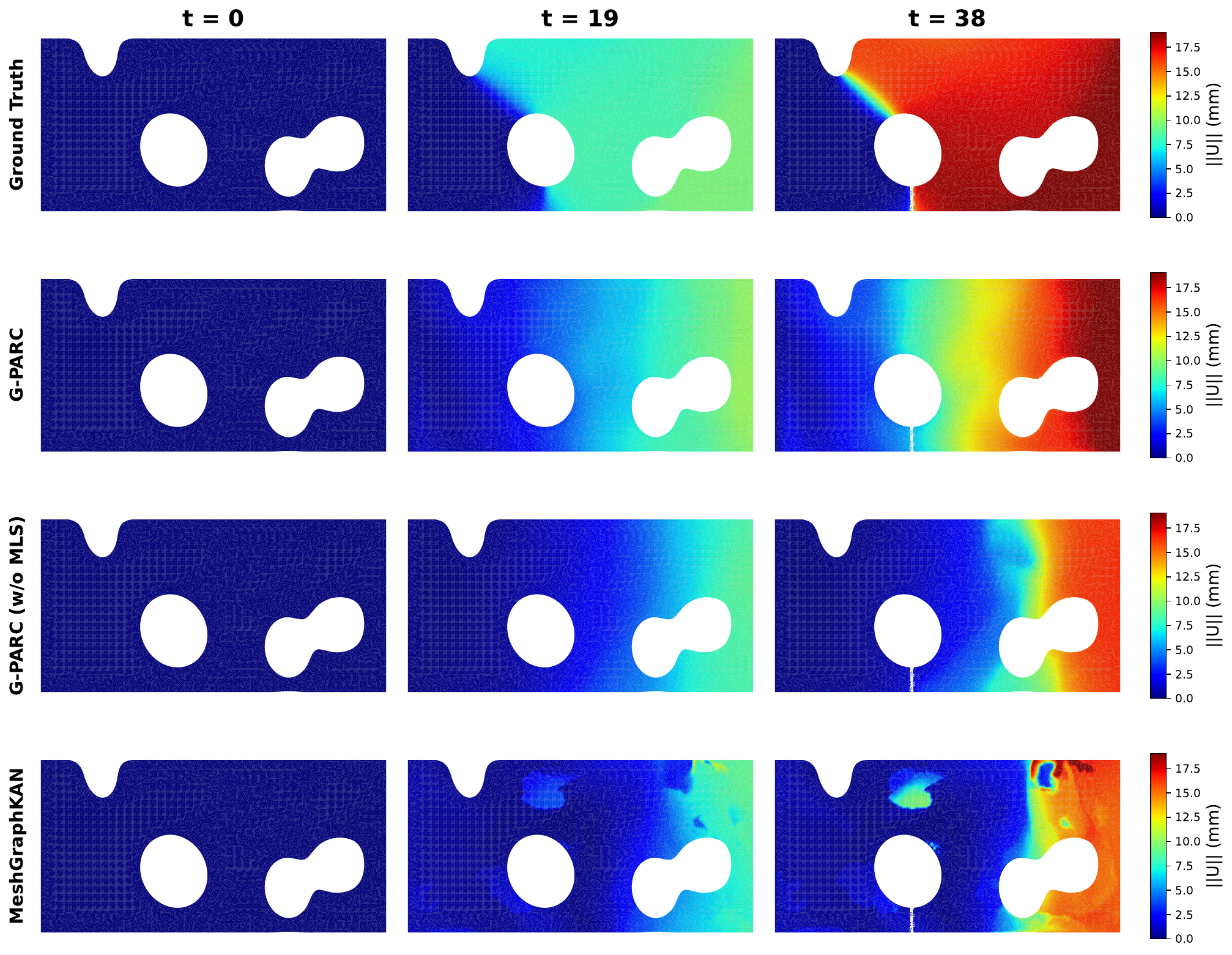}
    \caption[\textbf{Top three best performing models for elastoplastic dataset: G-PARC, G-PARC (\textbf{w/o MLS}), and MGKAN.}]{\textbf{Top three best performing models for elastoplastic dataset: G-PARC, G-PARC (\textbf{w/o MLS}), and MGKAN.}}
    \label{fig:elastoplasto}
\end{figure}

Table~\ref{tab:elasto} reports performance across all models on the elastoplastic dataset and Figure~\ref{fig:elastoplasto} depicts the top three performing models. G-PARC achieves the strongest performance on every metric, with an RRMSE AUC of 0.439, final-timestep RRMSE of 0.461, and SSIM AUC of 0.435. Unlike the previous two datasets, the second-best model is the ablated variant G-PARC (\textbf{w/o MLS}). The full model reduces RRMSE AUC by 6.9\% and final-timestep RRMSE by 12.1\% relative to the ablation. Among the encoder-processor-decoder baselines, MGKAN again emerges as the most competitive, though the gap to G-PARC is considerably wider here than in the fluvial hydrology and planar shock wave domains. The large standard deviations across all models reflect the inherent diversity of the test set, with each simulation featuring a unique mesh topology and node count. Critically, G-PARC consistently exhibits the lowest variance, suggesting that its physics-aware operators provide more robust generalization across geometrically diverse samples.

Across all three benchmarks, G-PARC consistently achieves the lowest errors and highest structural similarity, outperforming both its own ablated version  (G-PARC \textbf{w/o MLS}) and all encoder-processor-decoder baselines. The MLS ablation confirms that physics-aware differential operators are critical for approximating governing PDEs where standard message passing cannot approximate through learned weights alone. Notably, the performance of G-PARC (\textbf{w/o MLS}) and MGKAN shifts across domains. MGKAN is more capable of capturing advection and diffusion phenomena present in both fluvial hydrology and planar shock tube benchmarks. However, G-PARC (\textbf{w/o MLS}) outperforms MGKAN on the structural mechanics problem where topology significantly varies. The standard GNN architectures (MGNET and GraphSAGE) are unviable across all three benchmarks, suggesting that standard message-passing alone is insufficient for problems involving discontinuities, large deformations, evolving mesh topology. These results motivate the question of computational cost: if G-PARC’s accuracy gains came at prohibitive expense, their practical value would be limited and scaling to large systems infeasible. We next examine the throughput and scaling characteristics of each model. 

\subsection{Computational Performance}
Measured on a A100-SXM4-80GB GPU, G-PARC achieves the highest prediction accuracy across all three benchmark problems while requiring 2-3x fewer parameters and 3-6x fewer FLOPs per step than encoder-decoder baselines (\textbf{Table~\ref{tab:compute}}).  Node·steps/s normalizes the model throughput by mesh size, computed as (mesh nodes per simulation) x (autoregressive steps per rollout) / (wall-clock rollout time in seconds). One step is a single forward pass advancing the simulation one timestep. On the largest benchmark (elastoplastic, 25k to 32k nodes, 40 steps), G-PARC achieves a throughput of 4.05M node-steps/s. Compared to 6.7K for GINO and 1.7K for GNO, representing $604\times$ and $2{,}383\times$ throughput advantages respectively. This bottleneck is inherent to the kernel integral transform in GNO, which requires a neighborhood search every forward step: GNO requires 14.9 s per step on the elastoplastic mesh (vs.\ 6.5 ms for G-PARC), while consuming 8.3 GB of GPU memory for a single inference pass. Training GNO and GINO was not feasible within A100-80GB memory constraints; both were evaluated only for computational cost and excluded from accuracy comparisons.

\begin{table}[ht]
\centering

\resizebox{\textwidth}{!}{%
\begin{tabular}{lcccccc}
\toprule
Model & Params & Step (ms) & Rollout (s) & Node$\cdot$steps/s & GPU (MB) & GFLOPs/step \\
\midrule
\multicolumn{7}{l}{\textit{Fluvial Hydrology (2400-4800 nodes, 21--112 steps, variable)}} \\
\midrule
G-PARC    & 224K & \textbf{5.79 $\pm$ 0.13} & 0.241 & 0.75M & \textbf{704}  & \textbf{2.72}  \\
MGKAN     & 679K & 3.54 $\pm$ 0.01          & 0.148 & 1.22M & 1246          & 16.82          \\
MGNET     & 516K & 2.78 $\pm$ 0.01          & 0.116 & 1.56M & 1123          & 12.91          \\
GraphSAGE & 198K & 4.36 $\pm$ 0.01          & 0.181 & 0.99M & 772           & 3.48           \\
GNO       & 250K & 2844                     & 142.2 & 1.7K  & 778           & ---            \\
GINO      & 230K & 852                      & 42.6  & 5.6K  & 101           & ---            \\
\midrule
\multicolumn{7}{l}{\textit{Shock Tube (4,096 nodes, 42 steps)}} \\
\midrule
G-PARC    & 249K & \textbf{4.32 $\pm$ 0.06} & 0.173 & 0.95M & \textbf{220}  & \textbf{2.30}  \\
MGKAN     & 673K & 3.21 $\pm$ 0.01          & 0.128 & 1.28M & 277           & 14.15          \\
MGNET     & 631K & 3.14 $\pm$ 0.01          & 0.125 & 1.31M & 264           & 13.42          \\
GraphSAGE & 602K & 3.71 $\pm$ 0.01          & 0.148 & 1.10M & 218           & 4.90           \\
GNO       & 249K & 2415                     & 101.4 & 1.7K  & 347           & ---            \\
GINO      & 229K & 770                      & 32.4  & 5.3K  & 194           & ---            \\
\midrule
\multicolumn{7}{l}{\textit{Elastoplastic (25k-32k nodes, 40 steps)}} \\
\midrule
G-PARC    & \textbf{285K} & \textbf{6.50 $\pm$ 0.28} & \textbf{0.254} & \textbf{4.05M} & 528  & \textbf{34.67}  \\
MGKAN     & 667K          & 9.90 $\pm$ 0.49          & 0.386          & 2.66M          & 1373 & 122.85          \\
MGNET     & 514K          & 8.74 $\pm$ 0.43          & 0.341          & 3.01M          & 1287 & 94.87           \\
GraphSAGE & 498K          & 12.15 $\pm$ 0.62         & 0.474          & 2.17M          & 364  & 61.16           \\
GNO       & 248K          & 14921                    & 589.9          & 1.7K           & 8265 & ---             \\
GINO      & 228K          & 3774                     & 147.2          & 6.7K           & 347  & ---             \\
\bottomrule
\end{tabular}
}%
\caption{\textbf{Computational efficiency on NVIDIA A100-SXM4-80GB.} Step time = mean $\pm$ std (ms) per autoregressive step averaged over 200 rollouts (20 runs $\times$ 10 simulations). Rollout = mean wall-clock time for one complete simulation. Node$\cdot$steps/s normalizes throughput by mesh size. GFLOPs/step estimated via \texttt{torch.profiler} single-step measurement. For G-PARC, MLS operator initialization is amortized into the per-step cost (one-time per mesh topology). Best value per metric highlighted in green.}
\label{tab:compute}
\end{table}

\section{Discussion}

This work is the first to extend PARC from Cartesian grids to unstructured meshes by embedding analytically computed differential operators into graph message passing via MLS. Across three physics domains of increasing complexity, G-PARC demonstrates that explicit physics encoding is not only interpretable but necessary for stable long-horizon prediction. MGNET and GraphSAGE, which must implicitly discover these differential relationships through data-driven message passing catastrophically diverge during autoregressive rollout on all three benchmarks. MGKAN maintains stability but requires 2.4--3$\times$ more parameters for comparable or worse accuracy. G-PARC achieves the best RRMSE on every dataset with 224k-285k parameters, including near-exact shock reconstruction (RRMSE $= 0.0070$, $R^2 = 0.9998$)---an $8\times$ improvement over the next-best model---and the strongest performance on the 25--32k-node elastoplastic benchmark where all data-driven baselines fail.

The computational efficiency of G-PARC further distinguishes it from existing approaches. Because MLS operators analytically provide the differential quantities that other networks must learn from data, G-PARC requires 3--6$\times$ fewer FLOPs per step and exhibits efficient scaling with larger datasets. For the Elastoplastic dataset, the number node·steps/s increases only 1.5x for a 6.1x increase in node count, compared to $2.8$--$3.3\times$ for the baseline message passing encoder-processor-decoder architectures. This efficiency is not incidental but an advantage of explicitly incorporating numerical methods into neural network design. G-PARC results in a robust interpretable neural architecture scalable for real world unstructured nonlinear applications.

\section{Methods}

\subsection{G-PARC Architecture}

According to Chen et al., neural networks can approximate any nonlinear operator, establishing a foundation for operator learning with neural networks \cite{chen1995universal}. Following this theoretical foundation, PARC uses convolutional neural networks to learn a differential operator that is then used to predict a solution in a manner analogous to finite difference schemes. While this formulation has been well established on Euclidean data, defining differential operators in G-PARC on irregular graph structures using a message-passing framework is non-trivial.

In G-PARC, we consider a broad class of time-dependent PDEs that govern the spatiotemporal evolution of a state field $\mathbf{s}$ over a discrete spatial domain $\boldsymbol{\Omega} \subset \mathbb{R}^{d}$. In the most general form, the temporal evolution is determined by a nonlinear PDE of the state and its spatial derivatives:

\begin{equation}
    \frac{\partial \mathbf{s}(\mathbf{r},\, t)}{\partial t} = F\!\left(\mathbf{s},\, \nabla \mathbf{s},\, \Delta \mathbf{s}\right) + R(\mathbf{s},\, \mathbf{c})
    \label{eq:pde}
\end{equation}

Here, $\mathbf{s}(\mathbf{r}, t)$ denotes the state at position $\mathbf{r} \in \Omega$ at time $t$. The operators $\nabla$ and $\Delta = \nabla^{2}$ denote the gradient and Laplacian, respectively. $F$ denotes the general PDE operator (e.g., flux) and $R$ denotes the source term. This general form encompasses a wide range of physical phenomena. For example, setting $F \coloneqq k\Delta \mathbf{s} - \mathbf{u}\,\nabla \mathbf{s} + R$ recovers the advection-diffusion-reaction equation, the structure implemented in PARCv2. Similarly, the compressible Euler equations, elastodynamic equations, and shallow water equations can each be derived as specific instantiations with their own state variables, flux functions, and source terms. G-PARC approximates this entire nonlinear PDE with a single learned operator:
\begin{equation}
    \Phi_{\theta}\!\left(\mathbf{s},\, \nabla \mathbf{s},\, \Delta \mathbf{s},\, \mathbf{c}\right) = F\!\left(\mathbf{s},\, \nabla \mathbf{s},\, \Delta \mathbf{s}\right) + R\!\left(\mathbf{s},\, \mathbf{c}\right)
    \label{eq:operator}
\end{equation}

First, the differential operators $\nabla$ and $\nabla^2$ are approximated with MLS for the current time-step.

\begin{figure}
    \centering
    \includegraphics[width=\linewidth]{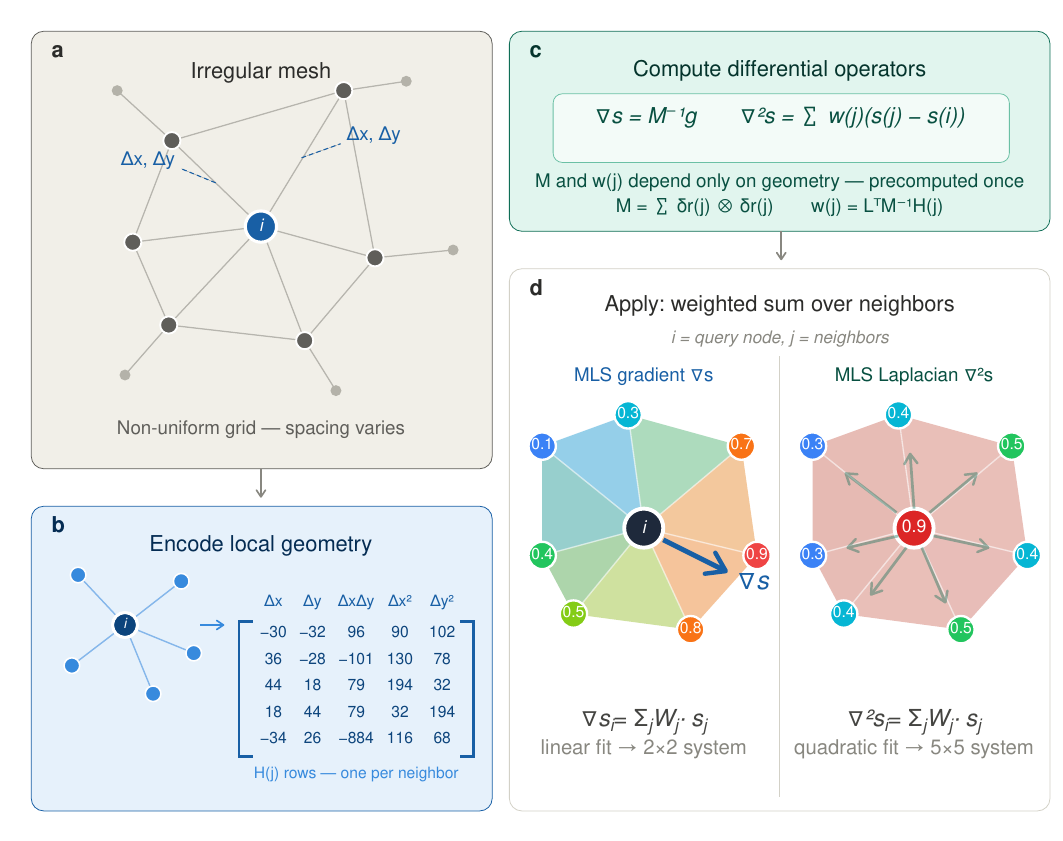}
    \caption[\textbf{Overview of the MLS differential operator construction used in G-PARC.}]{\textbf{Overview of the MLS differential operator construction used in G-PARC.} \textbf{(a)} An irregular mesh with non-uniform node spacing, where classical finite difference stencils are invalid. \textbf{(b)} For each node $v_i$, the relative position of each neighbor $v_j$ is encoded as a quadratic basis vector $\mathbf{H}_{ij}$, assembling one row per neighbor. \textbf{(c)} Differential operators are computed directly from local geometry. \textbf{(d)} The resulting operators are applied as weighted sums over the graph neighborhood $\mathcal{N}_i$ to produce the MLS gradient $\nabla s_i$ and Laplacian $\nabla^2 s_i$ at each node, which are embedded directly in the G-PARC computational graph.}
    \label{fig:mlsdiagram}
\end{figure}

We denote the graph $\mathcal{G} = (\mathbf{V},\, \mathcal{E})$, where $\mathbf{V} = \{v_i\}_{i=1}^{n}$ enumerates the $n$ graph nodes $v_i$, and $\mathcal{E}$ is the set of edges $e_{ij} = (v_i,\, v_j)$ where $i,j \in \{1,\ldots,n\}$. Each data sample has fixed topology across time, though topology can vary across different samples. The state field is discretized at each node $v_i$ at time $t$ and is denoted $\mathbf{s}(v, t)$. To approximate the differential operators $\nabla$ and $\nabla^{2}$ on a graph $\mathcal{G}$ with irregular node positions and connectivity, we use an MLS approximation that constructs local adaptive differential operators from the neighborhood of each node. This is done by fitting local polynomial expansions to the neighborhood and then analytically differentiating the resulting polynomials. Zhang et al.~\cite{zhang2025combining} provide a detailed derivation of a least-squares-based approximation of differential operators on graph structures, which we follow in this paper. However, while the MLS framework in~\cite{zhang2025combining} employs these operators in the loss function, G-PARC instead embeds them directly as architectural components. A diagram describing how MLS is used to derive spatial derivatives is shown in Figure~\ref{fig:mlsdiagram}.

Using MLS, the local gradient of $\mathbf{s}$ at node $v_i$ is approximated by fitting a linear polynomial to the field values across its neighborhood $\mathcal{N}_{v_i}$. The MLS gradient at each node, denoted $\nabla \mathbf{s}_i$, is computed by solving the normal equations of a local linear regression on each neighborhood $\mathcal{N}_{v_i}$:

\begin{equation}
    \nabla \mathbf{s}_i = \mathbf{M}_i^{-1}\, \mathbf{g}_i
    \label{eq:gradient}
\end{equation}

Here, $\mathbf{M}_i$ is a $2 \times 2$ moment matrix for the gradient defined by Equation~\ref{eq:moment}, where $\otimes$ denotes the outer product. Since $\mathbf{M}_i$ depends only on mesh geometry, it is precomputed once per mesh configuration and cached for computational efficiency. The vector $\mathbf{g}_i$ is assembled from the neighborhood as shown in Equation~\ref{eq:g}, and captures how the state changes along each edge. For each edge $e_{ij} \coloneqq (v_i,\, v_j) \in \mathcal{E}$, we define $\delta\mathbf{r}_{ij} = v_j - v_i$ as the relative displacement vector and $\delta\mathbf{s}_{ij} = \mathbf{s}_j - \mathbf{s}_i$ as the difference between neighboring states.

\begin{equation}
    \mathbf{M}_i = \sum_{j \in \mathcal{N}_{v_i}} \delta\mathbf{r}_{ij} \otimes \delta\mathbf{r}_{ij}
    \label{eq:moment}
\end{equation}

\begin{equation}
    \mathbf{g}_i = \sum_{j \in \mathcal{N}_{v_i}} \delta\mathbf{r}_{ij} \cdot \delta\mathbf{s}_{ij}
    \label{eq:g}
\end{equation}

The Laplacian of $\mathbf{s}$ is approximated with a quadratic polynomial basis. The relative position $\delta\mathbf{r}_{ij}$ is mapped to a five-term basis vector for each edge $e_{ij}$, where $x_{ij}$ and $y_{ij}$ denote the directional components of each edge:

\begin{equation}
    \mathbf{H}_{ij} = \left[\delta x_{ij},\; \delta y_{ij},\; \delta x_{ij}^{2},\; \delta y_{ij}^{2},\; \delta x_{ij} \cdot \delta y_{ij}\right]^{\mathsf{T}}
    \label{eq:basis}
\end{equation}

$\mathbf{H}_{ij}$ is then used to define $\mathbf{\tilde{M}}_i$, a distinct $5 \times 5$ moment matrix:

\begin{equation}
    \mathbf{\tilde{M}}_{i} = \sum_{j \in \mathcal{N}_i} \mathbf{H}_{ij}\mathbf{H}_{ij}^{\mathsf{T}}
    \label{eq:moment5}
\end{equation}

Applying the Laplacian to the quadratic basis gives a constant vector,
\begin{equation}
    \mathbf{L} := \nabla^2 \mathbf{H} = [0, 0, 2, 2, 0]^T,
\end{equation}
so the Laplacian $\nabla^2 s$ reduces to a weighted sum over edges:
\begin{equation}
    \nabla^2 \mathbf{s} = \sum_{j \in \mathcal{N}_i} w_{ij}(\mathbf{s}_j - \mathbf{s}_i)
\end{equation}
\begin{equation}
    w_{ij} = \mathbf{L}^T \tilde{\mathbf{M}}_i^{-1} \mathbf{H}_{ij}
\end{equation}

The weights $w_{ij}$ depend only on mesh geometry and are precomputed and cached at the start of model training.

The operators $\nabla\mathbf{s}$, $\nabla^{2}\mathbf{s}$, and $\mathbf{s}$ are concatenated with the source term $R$, which is approximated with a GNN:

\begin{equation}
    R(\mathbf{s},\, \mathbf{c}) \approx R_{\theta_R}(\mathbf{s},\, \mathbf{c})
    \label{eq:source}
\end{equation}

And fed through a final MLP:

\begin{equation}
    \frac{\partial \mathbf{s}}{\partial t} \approx \Phi_{\theta}\!\left(\mathbf{s},\, \nabla\mathbf{s},\, \nabla^{2}\mathbf{s},\, \mathbf{c}\right) \coloneqq \mathrm{MLP}_{\theta_{\mathrm{MLP}}}\!\left(\left[\mathbf{s},\, \nabla\mathbf{s},\, \nabla^{2}\mathbf{s}\right] \,\|\, R_{\theta_R}(\mathbf{s},\, \mathbf{c})\right)
    \label{eq:mlp}
\end{equation}

Once $\Phi_{\theta}$ is learned, it can be used to predict $\frac{\partial \mathbf{s}}{\partial t}$ and advance the state $\mathbf{s}$ forward in time. The initial value problem of Equation~\ref{eq:pde}, subject to the initial condition $\mathbf{s}(t=0) = \mathbf{s}_0$, is approximated as:

\begin{equation}
    \mathbf{s}_{t+\Delta t} = \mathbf{s}_t + \Delta\mathbf{s}, \qquad \Delta\mathbf{s} = \int_{0}^{\Delta t} \Phi_{\theta}\!\left(\mathbf{s}_t,\, \nabla\mathbf{s}_t,\, \nabla^{2}\mathbf{s}_t,\, \mathbf{c}\right)\, dt
    \label{eq:integration}
\end{equation}

Here, $\mathbf{s}_t$ denotes the state at time $t$ and $\Delta t$ is the desired step forward in time. The integral is evaluated numerically. In this work we use the forward Euler method, though higher-order schemes such as Runge-Kutta and Heun's method can also be used.

The training objective is to minimize the discrepancy between the predicted next state and the ground truth. Because $\Phi_{\theta}$ outputs a derivative that the integrator applies to the current state, the loss is computed on full states rather than increments. Using the $L^2$ loss:

\begin{equation}
    \mathcal{L}(\theta) = \frac{1}{N}\sum_{i=1}^{N}\left\|\hat{s}_{t+\Delta t}^{(i)} - s_{t+\Delta t}^{(i),\,\mathrm{true}}\right\|_{2}^{2}
    \label{eq:loss_single}
\end{equation}

where $\hat{s}_{t+\Delta t} = s_t + \Delta s$ is the predicted state, $s_{t+\Delta t}^{\mathrm{true}}$ is the ground-truth state, and $N$ is the number of nodes. For multi-step sequences, the loss is averaged over timesteps:

\begin{equation}
    \mathcal{L}(\theta) = \frac{1}{NT}\sum_{t=0}^{T-1}\sum_{i=1}^{N}\left\|\hat{s}_{t+\Delta t}^{(i)} - s_{t+\Delta t}^{(i),\,\mathrm{true}}\right\|_{2}^{2}
    \label{eq:loss_multi}
\end{equation}

allowing backpropagation through the numerical integration of $\Phi_{\theta}$.

\subsection*{Training Protocol}

All models were trained under the same autoregressive scheme to ensure fair comparison across datasets. G-PARC was optimized using AdamW with a cosine annealing learning rate schedule ($\eta_{\min} = \eta_0 \times 0.01$) over the full training duration. The number of epochs is set per dataset based on convergence behavior: 100 epochs for the shock tube domain ($\mathrm{LR} = 3\times10^{-4}$), 200 epochs for the fluvial hydrodynamics domain ($\mathrm{LR} = 1\times10^{-4}$), and 1{,}500 epochs for the elastoplastic domain ($\mathrm{LR} = 3\times10^{-4}$). In all cases, the best-performing checkpoint on the validation set was retained for evaluation.

All models were trained with a pure free-running autoregressive rollout strategy, whereby the model receives its own predictions as inputs across a sequence of $K$ steps during each training iteration, exactly matching the inference-time prediction strategy. Sequence lengths of $K = 16$ steps were used for the elastoplastic domain and $K = 4$ steps for both the shock tube and fluvial hydrodynamics domains, with non-overlapping strides matching the sequence length.

The MGKAN and MGNET baselines were adopted from NVIDIA's PhysicsNeMo library~\cite{physicsnemo2024} and trained using the suggested default learning rates and hyperparameter configurations. No additional architecture-specific tuning was performed beyond what is described here, ensuring that observed performance differences reflect model capacity and inductive bias rather than optimization advantage.

\section*{Data Availability}

The implementation of this work is available on GitHub (\url{https://github.com/JackBeerman/G-PARC/tree/main}). Data utilized in this study may be obtained from the authors upon reasonable request.

\section*{Acknowledgments}

This work was supported by the U.S. Department of Energy under the National Nuclear Security Administration (NNSA) Stewardship Science Academic Alliances (SSAA) Program (Grant No. DE-NA0004239), and partially by the PSAAP IV SAGEST Center at the University of Virginia, under contract number DE-NA0004269.

\section*{Author Contributions} \textbf{Jack T. Beerman}: Conceptualization, Methodology, Software, Writing – original draft, Writing – review \& editing. \textbf{Tyler J. Abele}: Software, Writing – review \& editing. \textbf{Mehdi Taghizadeh}: Data curation, Writing – original draft, Writing – review \& editing. \textbf{Andrew Davis}: Data curation, Writing – original draft, Writing – review \& editing. \textbf{Zoe J. Gray}: Validation, Writing – review \& editing. \textbf{Negin Alemazkoor}: Supervision, Writing – review \& editing. \textbf{Xifeng Gao}: Supervision, Writing – review \& editing. \textbf{H.S. Udaykumar}: Writing – review \& editing. \textbf{Stephen S. Baek}: Conceptualization, Funding acquisition, Investigation, Project administration, Supervision, Writing – review \& editing.

\bibliographystyle{unsrt} 
\bibliography{ref} 

@article{zhang2025combining,
  title={Combining physics-informed graph neural network and finite difference for solving forward and inverse spatiotemporal PDEs},
  author={Zhang, Hao and Jiang, Longxiang and Chu, Xinkun and Wen, Yong and Li, Luxiong and Liu, Jianbo and Xiao, Yonghao and Wang, Liyuan},
  journal={Computer Physics Communications},
  volume={308},
  pages={109462},
  year={2025},
  publisher={Elsevier}
}

@article{nguyen2024parcv2,
  title={PARCv2: Physics-aware recurrent convolutional neural networks for spatiotemporal dynamics modeling},
  author={Nguyen, Phong CH and Cheng, Xinlun and Azarfar, Shahab and Seshadri, Pradeep and Nguyen, Yen T and Kim, Munho and Choi, Sanghun and Udaykumar, HS and Baek, Stephen},
  journal={arXiv preprint arXiv:2402.12503},
  year={2024}
}

@article{nguyen2023parc,
  title={PARC: Physics-aware recurrent convolutional neural networks to assimilate meso scale reactive mechanics of energetic materials},
  author={Nguyen, Phong CH and Nguyen, Yen-Thi and Choi, Joseph B and Seshadri, Pradeep K and Udaykumar, HS and Baek, Stephen S},
  journal={Science advances},
  volume={9},
  number={17},
  pages={eadd6868},
  year={2023},
  publisher={American Association for the Advancement of Science}
}

@article{li2020fourier,
  title={Fourier neural operator for parametric partial differential equations},
  author={Li, Zongyi and Kovachki, Nikola and Azizzadenesheli, Kamyar and Liu, Burigede and Bhattacharya, Kaushik and Stuart, Andrew and Anandkumar, Anima},
  journal={arXiv preprint arXiv:2010.08895},
  year={2020}
}

@article{li2023geometry,
  title={Geometry-informed neural operator for large-scale 3d pdes},
  author={Li, Zongyi and Kovachki, Nikola and Choy, Chris and Li, Boyi and Kossaifi, Jean and Otta, Shourya and Nabian, Mohammad Amin and Stadler, Maximilian and Hundt, Christian and Azizzadenesheli, Kamyar and others},
  journal={Advances in Neural Information Processing Systems},
  volume={36},
  pages={35836--35854},
  year={2023}
}

@inproceedings{NEURIPS2021_df438e52,
 author = {Krishnapriyan, Aditi and Gholami, Amir and Zhe, Shandian and Kirby, Robert and Mahoney, Michael W},
 booktitle = {Advances in Neural Information Processing Systems},
 editor = {M. Ranzato and A. Beygelzimer and Y. Dauphin and P.S. Liang and J. Wortman Vaughan},
 pages = {26548--26560},
 publisher = {Curran Associates, Inc.},
 title = {Characterizing possible failure modes in physics-informed neural networks},
 url = {https://proceedings.neurips.cc/paper_files/paper/2021/file/df438e5206f31600e6ae4af72f2725f1-Paper.pdf},
 volume = {34},
 year = {2021}
}

@inproceedings{gao2016high,
  title={A high-order finite-volume method for combustion},
  author={Gao, Xinfeng and Owen, Landon D and Guzik, Stephen M},
  booktitle={54th AIAA Aerospace Sciences Meeting},
  pages={1808},
  year={2016}
}

@article{wang2021understanding,
  title={Understanding and mitigating gradient flow pathologies in physics-informed neural networks},
  author={Wang, Sifan and Teng, Yujun and Perdikaris, Paris},
  journal={SIAM Journal on Scientific Computing},
  volume={43},
  number={5},
  pages={A3055--A3081},
  year={2021},
  publisher={SIAM}
}

@article{bartolucci2023neural,
  title={Are neural operators really neural operators? frame theory meets operator learning},
  author={Bartolucci, Francesca and de B{\'e}zenac, Emmanuel and Raoni{\'c}, Bogdan and Molinaro, Roberto and Mishra, Siddhartha and Alaifari, Rima},
  journal={SAM Research Report},
  volume={2023},
  year={2023},
  publisher={ETH Zurich}
}

@inproceedings{fanaskov2023spectral,
  title={Spectral neural operators},
  author={Fanaskov, Vladimir Sergeevich and Oseledets, Ivan V},
  booktitle={Doklady Mathematics},
  volume={108},
  pages={S226--S232},
  year={2023},
  organization={Springer}
}

@article{cheng2024physics,
  title={Physics-aware recurrent convolutional neural networks for modeling multiphase compressible flows},
  author={Cheng, Xinlun and Nguyen, Phong CH and Seshadri, Pradeep K and Verma, Mayank and Gray, Zo{\"e} J and Beerman, Jack T and Udaykumar, HS and Baek, Stephen S},
  journal={International Journal of Multiphase Flow},
  volume={177},
  pages={104877},
  year={2024},
  publisher={Elsevier}
}

@article{chen1995universal,
  title={Universal approximation to nonlinear operators by neural networks with arbitrary activation functions and its application to dynamical systems},
  author={Chen, Tianping and Chen, Hong},
  journal={IEEE transactions on neural networks},
  volume={6},
  number={4},
  pages={911--917},
  year={1995},
  publisher={IEEE}
}

@article{cheng2025multi,
  title={Multi-resolution physics-aware recurrent convolutional neural network for complex flows},
  author={Cheng, Xinlun and Choi, Joseph and Udaykumar, HS and Baek, Stephen},
  journal={APL Machine Learning},
  volume={3},
  number={4},
  year={2025},
  publisher={AIP Publishing}
}

@article{gray2025reduced,
  title={Reduced Order Modeling of Energetic Materials Using Physics-Aware Recurrent Convolutional Neural Networks in a Latent Space (LatentPARC)},
  author={Gray, Zo{\"e} J and Choi, Joseph B and Choi, Youngsoo and Springer, H Keo and Udaykumar, HS and Baek, Stephen S},
  journal={arXiv preprint arXiv:2509.12401},
  year={2025}
}

@inproceedings{pfaff2020learning,
  title={Learning mesh-based simulation with graph networks},
  author={Pfaff, Tobias and Fortunato, Meire and Sanchez-Gonzalez, Alvaro and Battaglia, Peter},
  booktitle={International conference on learning representations},
  year={2020}
}

@misc{physicsnemo2024,
  author       = {{NVIDIA}},
  title        = {{NVIDIA PhysicsNeMo}: An Open-Source Framework for 
                  Physics-Informed Machine Learning},
  year         = {2024},
  howpublished = {\url{https://developer.nvidia.com/physicsnemo}},
  note         = {Accessed: March 2026}
}

@misc{openradioss2022,
  author       = {{OpenRadioss Community}},
  title        = {{OpenRadioss}: Open-Source Finite Element Solver for 
                  Dynamic Event Analysis},
  year         = {2022},
  howpublished = {\url{https://openradioss.org/}},
  note         = {Accessed: March 2026}
}

@article{beerman2026size,
  title={Size is Not the Solution: Deformable Convolutions for Effective Physics Aware Deep Learning},
  author={Beerman, Jack T and Roy, Shobhan and Udaykumar, HS and Baek, Stephen S},
  journal={arXiv preprint arXiv:2601.11657},
  year={2026}
}

@article{liu2024kan,
  title={Kan: Kolmogorov-arnold networks},
  author={Liu, Ziming and Wang, Yixuan and Vaidya, Sachin and Ruehle, Fabian and Halverson, James and Solja{\v{c}}i{\'c}, Marin and Hou, Thomas Y and Tegmark, Max},
  journal={arXiv preprint arXiv:2404.19756},
  year={2024}
}

@article{hamilton2017inductive,
  title={Inductive representation learning on large graphs},
  author={Hamilton, Will and Ying, Zhitao and Leskovec, Jure},
  journal={Advances in neural information processing systems},
  volume={30},
  year={2017}
}

@article{casenave2025physics,
  title={Physics-Learning AI Datamodel (PLAID) datasets: a collection of physics simulations for machine learning},
  author={Casenave, Fabien and Roynard, Xavier and Staber, Brian and Piat, William and Bucci, Michele Alessandro and Akkari, Nissrine and Kabalan, Abbas and Nguyen, Xuan Minh Vuong and Saverio, Luca and Perez, Rapha{\"e}l Carpintero and others},
  journal={arXiv preprint arXiv:2505.02974},
  year={2025}
}

@manual{HECRAS2025,
  author       = {{U.S. Army Corps of Engineers, Hydrologic Engineering Center}},
  title        = {HEC-RAS: River Analysis System},
  year         = {2025},
  note         = {Version 6.6},
  organization = {U.S. Army Corps of Engineers},
  address      = {Davis, CA},
  url          = {https://www.hec.usace.army.mil/software/hec-ras/}
}

@misc{USGS2023NWIS,
  author       = {{U.S. Geological Survey}},
  title        = {{USGS National Water Information System surface-water data}},
  year         = {2023},
  url          = {https://waterdata.usgs.gov},
  doi          = {10.5066/F7P55KJN}
}

@article{taghizadeh2025interpretable,
  title={Interpretable physics-informed graph neural networks for flood forecasting},
  author={Taghizadeh, Mehdi and Zandsalimi, Zanko and Nabian, Mohammad Amin and Shafiee-Jood, Majid and Alemazkoor, Negin},
  journal={Computer-Aided Civil and Infrastructure Engineering},
  year={2025},
  publisher={Wiley Online Library}
}

\appendix 
\section{Appendix}

\subsection*{Fluvial Hydrology Simulation Setup}
\label{app:fluvial}

The fluvial hydrology benchmark consists of spatiotemporal flood simulations generated on irregular HEC-RAS meshes over two real-world river-floodplain domains: the Iowa River in Marshall County, Iowa, and the White River in Delaware County, Indiana. These two case studies were selected to provide substantial geometric and hydraulic diversity within the same benchmark. The Iowa River case represents a broader rural floodplain, whereas the White River case represents a narrower and more topographically complex river corridor. As a result, the benchmark captures both relatively smooth river-floodplain inundation patterns and more heterogeneous flow responses shaped by irregular terrain and localized conveyance pathways.

The two case studies also differ substantially in spatial extent and mesh topology. Both domains are represented using irregular meshes with 80~m spatial resolution, but they differ in the number of mesh cells and connectivity structure. The Iowa River domain contains approximately 2,400 nodes and 9,900 edges and covers a river length of approximately 7.2~km with an average channel width of 60~m. The White River domain contains approximately 4,800 nodes and 19,200 edges; the modeled reach spans approximately 4.3~km with an average channel width of about 160~m. Together, these domains expose the model to substantial variation in mesh topology, node count, cell geometry, channel-floodplain connectivity, and local terrain gradients.

Each simulation is generated by running a two-dimensional HEC-RAS model on the corresponding river mesh under varied forcing conditions. For the Iowa River case, historical base hydrographs are used as templates and randomly scaled to create diverse inflow scenarios spanning lower- and higher-flow events. For the White River case, base hydrographs and rainfall hyetographs are scaled to create varied combinations of river inflow and rain-on-grid forcing. The resulting hydraulic fields are then post-processed into graph-structured temporal samples by extracting nodewise hydraulic states at successive output times.

The benchmark contains 1,000 simulation rollouts, partitioned into 700 training, 150 validation, and 150 test cases. The simulations have variable rollout lengths across cases, which further increases the temporal diversity of the dataset. Each mesh cell is treated as a graph node, and the evolving hydraulic state is represented through water depth, cell volume, and depth-averaged velocity components. As summarized in Table~\ref{tab:fluvial_features}, static terrain and geometric attributes are paired with dynamic hydraulic variables, making the dataset suitable for learning nonlinear spatiotemporal dynamics on unstructured meshes across heterogeneous river environments.

\begin{table}[!h]
  \centering
  \renewcommand{\arraystretch}{1.3}
  \resizebox{\textwidth}{!}{%
    \begin{tabular}{|l|p{6.5cm}|p{6.5cm}|}
      \hline
      \textbf{Feature (Symbol)} & \textbf{Description} & \textbf{Calculation} \\
      \hline\hline
      \multicolumn{3}{|l|}{\textbf{Static features}} \\ 
      \hline
      Coordinates $(x,y)$ & Planar centroid of each cell (m). 
                          & Average of all vertex $(x,y)$ coordinates. \\
      \hline
      Area $(A)$ & Surface area of each cell (m$^2$). 
                 & Polygon area based on cell vertices. \\
      \hline
      Elevation $z_b$ & Minimum bed elevation in each cell (m). 
                      & $\min$ of all subgrid elevation points within the cell. \\
      \hline
      Slope $S$ & Terrain slope within each cell (unitless). 
                & 
                \( S = \frac{\Delta z}{\Delta \ell},\;
                   \Delta \ell = \sqrt{(x_{\rm high}-x_{\rm low})^2 + (y_{\rm high}-y_{\rm low})^2} \). \newline (Approximated from subgrid elevations.) \\
      \hline
      Aspect $\phi$ & Flow‐direction aspect angle (°). 
                    & $\operatorname{atan2}(\Delta y,\Delta x)$ of steepest descent. \\
      \hline
      Curvature $\kappa$ & Planform curvature of terrain (m$^{-1}$). 
                          & Second derivative of subgrid elevation along cell centerline. \\
      \hline
      Manning’s $n$ & Channel roughness coefficient (unitless). 
                     & User-specified parameter (commonly expressed in s/m$^{1/3}$ in mixed-unit systems). \\
      \hline
      Flow accumulation (FA) & Upslope contributing area (m$^2$). 
                             & Sum of upslope cell areas based on DEM-derived flow paths. \\
      \hline\hline
      \multicolumn{3}{|l|}{\textbf{Dynamic features}} \\ 
      \hline
      Water depth $h$ & Depth of water column above bed (m). 
                       & $h(t)=\eta(t)-z_b$ (using subgrid bed elevation). \\
      \hline
      Volume $V$ & Storage volume in each cell (m$^3$). 
                  & Derived from subgrid elevation–area–volume relationships within the cell. \\
      \hline
      Velocity $u$ (x‐dir) & Depth‐averaged velocity in X (m/s). 
                           & SWE‑ELM output “Velocity~X.” \\
      \hline
      Velocity $v$ (y‐dir) & Depth‑averaged velocity in Y (m/s). 
                           & SWE‑ELM output “Velocity~Y.” \\
      \hline
    \end{tabular}%
  }
  \caption{Static and dynamic features extracted from each HEC‑RAS mesh cell, incorporating subgrid bathymetry.}
  \label{tab:fluvial_features}
\end{table}

The governing equations are the depth-averaged shallow water equations, where $h$ denotes water depth, $u$ and $v$ are the depth-averaged velocity components in the $x$- and $y$-directions, $z_b$ denotes bed elevation, and $\eta = h + z_b$ denotes the water-surface elevation. The conservation of mass is written as

\begin{equation}
    \frac{\partial h}{\partial t} + \frac{\partial(hu)}{\partial x} + \frac{\partial(hv)}{\partial y} = q,
    \label{eq:swe_mass}
\end{equation}

where $q$ denotes an external source or sink term when present. The momentum equations in the two horizontal directions are

\begin{align}
    \frac{\partial u}{\partial t} + u\frac{\partial u}{\partial x} + v\frac{\partial u}{\partial y} &= -g\frac{\partial\eta}{\partial x} + \nu_t\!\left(\frac{\partial^2 u}{\partial x^2} + \frac{\partial^2 u}{\partial y^2}\right) - gn^2\frac{u\sqrt{u^2+v^2}}{h^{4/3}}, \label{eq:swe_mom_u} \\
    \frac{\partial v}{\partial t} + u\frac{\partial v}{\partial x} + v\frac{\partial v}{\partial y} &= -g\frac{\partial\eta}{\partial y} + \nu_t\!\left(\frac{\partial^2 v}{\partial x^2} + \frac{\partial^2 v}{\partial y^2}\right) - gn^2\frac{v\sqrt{u^2+v^2}}{h^{4/3}}, \label{eq:swe_mom_v}
\end{align}

where $g$ is gravitational acceleration, $\nu_t$ is an effective eddy-viscosity coefficient, and $n$ is Manning's roughness coefficient. These equations describe the evolution of free-surface flow over irregular topography under the combined effects of advection, pressure gradients, diffusion, and bed friction. Water depth is obtained from

\begin{equation}
    h = \eta - z_b.
    \label{eq:water_depth}
\end{equation}

Cell volume is derived from the subgrid elevation-area-volume relationship within each HEC-RAS cell, and the velocity components are obtained from the depth-averaged hydraulic solution returned by the solver.

Because the benchmark combines two distinct river settings with substantially different mesh densities and terrain structure, it provides a challenging test of mesh-invariant generalization. Models must learn to propagate flow information across irregular cell neighborhoods while remaining robust to changes in domain extent, resolution, and hydraulic complexity. This makes the fluvial hydrology dataset an effective benchmark for evaluating surrogate models for unstructured spatiotemporal flood dynamics.

\section{Planar Shock Wave Simulation Setup}
\label{app:shockdata}

The shock tube problem is governed by the compressible Euler equations,
\begin{equation}
    \frac{\partial \mathbf{U}}{\partial t} + \nabla \cdot \vec{F}(\mathbf{U}) = 0
    \label{eq:euler_conservative}
\end{equation}
where
\begin{equation}
    \mathbf{U} = \begin{bmatrix} \rho \\ \rho\vec{u} \\ E \end{bmatrix}, \qquad
    \vec{F}(\mathbf{U}) = \begin{bmatrix} \rho\vec{u} \\ \rho\vec{u}\vec{u} + p\mathbf{I} \\ \vec{u}(E + p) \end{bmatrix}
    \label{eq:euler_flux}
\end{equation}

\begin{figure}[htbp]
    \centering
    \includegraphics[width=\linewidth]{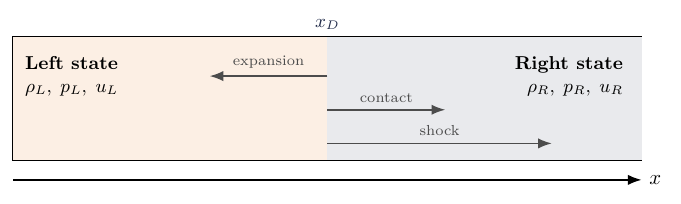}
    \caption[\textbf{Schematic of a typical planar shock wave problem.}]{\textbf{Schematic of a typical planar shock wave problem.} $x_D$ 
    denotes the location of the diaphragm. Rupture of the diaphragm 
    produces an expansion fan (left), a contact surface, and a shock 
    wave (right).}
    \label{fig:shocktube_schematic}
\end{figure}

Here $\rho$ (kg/m$^3$) is density, $\vec{u} = (u_x, u_y)^\top$ (m/s) is the velocity vector, $E$ (kg/m/s$^2$) is the total energy per unit volume, and $p$ (Pa) is the pressure. The system is closed with the ideal gas law $p = \rho R T$. The shock tube problem is a classical benchmark for verifying CFD solvers, combining an analytic solution with strong discontinuities that stress-test numerical algorithms. Figure~\ref{fig:shocktube_schematic} shows a shock tube initially consisting of two regions of gas at different pressures separated by a diaphragm, where the left state contains high pressure and density and the right state contains low pressure and density. Rupture of the diaphragm initiates three distinct flow features: an expansion fan that propagates into the left state lowering pressure and density isentropically; a contact surface with a physical density discontinuity but continuous pressure and velocity; and a shock wave traveling into the right state causing increased pressure and density across a strong discontinuity. Flow in the $y$-direction is uniform due to the absence of a pressure gradient.

The numerical solution is computed with the in-house CFD solver Chord~\cite{gao2016high}, a structured-grid fourth-order finite volume method. The finite volume method solves the integral form of the compressible Euler equations:
\begin{equation}
    \frac{d\langle \mathbf{U} \rangle_i}{dt} = 
    -\frac{1}{\Delta x} \sum_{d=1}^{D} \left(
    \langle \vec{F} \rangle_{i + \frac{1}{2}e^d} - 
    \langle \vec{F} \rangle_{i - \frac{1}{2}e^d}
    \right)
    \label{eq:fvm_euler}
\end{equation}
where the subscript $i$ is a multidimensional cell index, $\frac{1}{2}e^d$ denotes neighboring face values, and angle brackets denote cell-averaged quantities. Face-averaged fluxes $\langle \vec{F} \rangle_{i+\frac{1}{2}e^d}$ are computed by reconstructing face-averaged primitive variables $W$ (mass, velocity, pressure) using a fourth-order stencil:
\begin{equation}
    \langle W \rangle_{i+\frac{1}{2}e^d} = 
    -\frac{1}{12}\langle W \rangle_{i-e^d} 
    + \frac{7}{12}\langle W \rangle_{i} 
    + \frac{7}{12}\langle W \rangle_{i+e^d} 
    - \frac{1}{12}\langle W \rangle_{i+2e^d} 
    + \mathcal{O}(\Delta x^4)
    \label{eq:stencil}
\end{equation}
Time integration uses the standard fourth-order Runge--Kutta scheme. In regions of discontinuity, a partial parabolic limiter reduces the stencil order to suppress spurious oscillations. All calculations use double floating-point precision and are written to HDF5 files.

The shock tube dataset comprises 500 simulation cases, each consisting of 43 consecutive timesteps, yielding 21,500 total samples. Cases are generated by varying the left-state pressure $p_L$ and density $\rho_L$ while maintaining fixed ratios $p_L / p_R = 10$ and $\rho_L / \rho_R = 8$. The pressure takes 20 uniformly spaced values in $p_L \in [50{,}000,\ 168{,}750]$ Pa with increments of 6,250 Pa, and the density takes 25 uniformly spaced values in $\rho_L \in [0.5,\ 2.0]$ kg/m$^3$ with increments of 0.0625 kg/m$^3$. Each case is advanced with a constant timestep $\Delta t$ determined by the CFL condition:
\begin{equation}
    \Delta t = \frac{0.5\,\Delta x}{C}
    \label{eq:cfl}
\end{equation}
where $C$ is the theoretical maximum wave speed. Lower densities and higher left-state pressures correspond to increased sound speed and therefore smaller $\Delta t$. Across the 500 cases, $\Delta t$ spans a $3.7\times$ range from $3.84 \times 10^{-6}$ s to $1.41 \times 10^{-5}$ s, with 381 unique values. The timestep $\Delta t$ is provided to the model as a global conditioning parameter alongside $p_L$ and $\rho_L$, enabling G-PARC to adapt its temporal dynamics to each case --- a critical distinction from PARCv2, which assumes uniform $\Delta t$ across all training samples.

\begin{figure}[!ht]
    \centering
    \includegraphics[width=\linewidth]{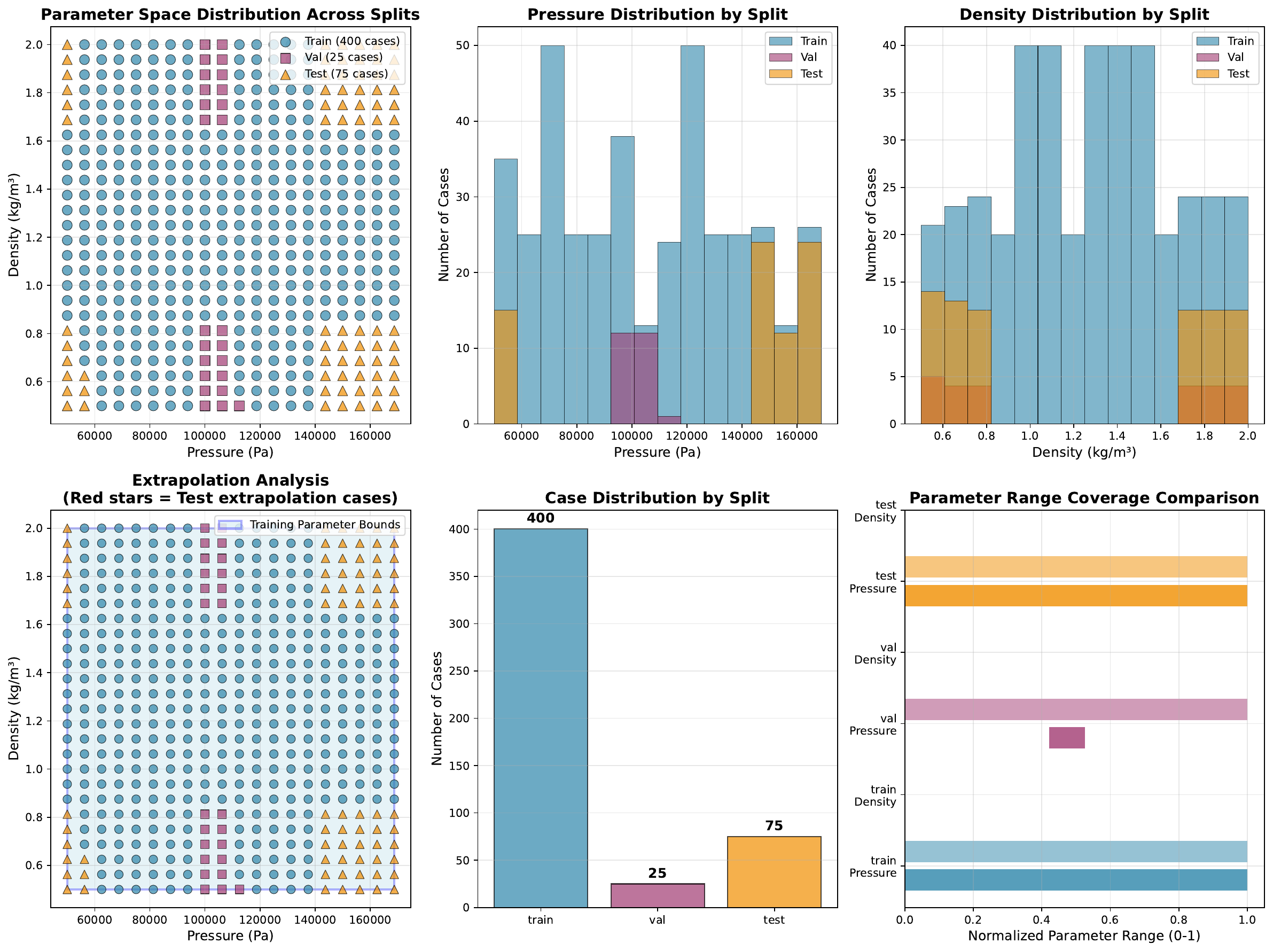}
    \caption[\textbf{Parameter space distribution across the 400 training, 25 validation, and 75 test cases in the $p_L$--$\rho_L$ domain.}]{\textbf{Parameter space distribution across the 400 training, 25 validation, and 75 test cases in the $p_L$--$\rho_L$ domain.} The split shows that training covers the full interior grid, validation is restricted to a narrow mid-range pressure band, and test cases are deliberately concentrated at the extremes to evaluate out-of-distribution 
extrapolation.}
    \label{fig:wavesplit}
\end{figure}

The dataset is partitioned as follows (\textbf{Figure~\ref{fig:wavesplit}}). The \textbf{training set} (400 cases) spans the full range of both $p_L$ and $\rho_L$, with all 20 pressure values and all 25 density values represented, providing broad coverage of the parameter space including the full range of $\Delta t$ values. The \textbf{validation set} (25 cases) is restricted to a narrow band of mid-range pressures ($p_L \in \{100{,}000,\ 106{,}250\}$ Pa, with one case at 112,500 Pa) paired with densities spanning the full range, evaluating interpolation performance at pressures well-represented during training. The \textbf{test set} (75 cases) is concentrated at the extremes of the parameter space, combining low pressures ($p_L \in \{50{,}000,\ 56{,}250\}$ Pa) and high pressures ($p_L \in \{143{,}750,\ 150{,}000,\ 156{,}250,\ 162{,}500,\ 168{,}750\}$ Pa) with extreme densities ($\rho_L \leq 0.8125$ kg/m$^3$ and $\rho_L \geq 1.6875$ kg/m$^3$). This split deliberately targets the corners of the parameter space, providing a stringent test of extrapolation beyond the densely sampled interior. Notably, the test set includes both the smallest and largest $\Delta t$ values in the dataset, further challenging the model's temporal generalization.

\section{Elastoplastic Simulation Setup}
\label{app:elastoplastic}

The elastoplastic dataset is derived from the PLAID benchmark suite using OpenRadioss~\cite{casenave2025physics}. The benchmark, denoted \texttt{2D\_ElPlDynamics}, consists of two-dimensional dynamic nonlinear structural mechanics simulations in large deformation and plane strain regimes, solved with the finite element method. The material is modeled with a nonlinear elastoplastic constitutive law incorporating damage via element erosion, failure, and a non-local regularization method for reducing mesh sensitivity. Each simulation computes the transient deformation of a 2D structure subjected to an imposed displacement on the right boundary and zero displacement on the left boundary~\cite{casenave2025physics}.

Input variability across the dataset arises from the unstructured meshes, which differ in shape, number of nodes, connectivity, and topology. The output fields of interest are the displacement components $U_x$ and $U_y$ all defined at the mesh nodes over the simulation horizon.

The full dataset contains 1,000 simulation cases. We randomly partition these into 50 training, 10 validation, and 932 test samples. Prior work utilized 40 A100 GPUs for training on the full dataset. We employ this small-data regime deliberately to evaluate model performance under limited supervision on a single A100-80GB GPU; better results across all models would be expected with access to the full training set.

\section{Evaluation Metrics}

\paragraph{Root Mean Squared Error (RMSE).} The RMSE at timestep $t$ measures the average magnitude of prediction error across all $N$ nodes and $D$ variables:
\begin{equation}
\mathrm{RMSE}(t) = \sqrt{\frac{1}{N} \sum_{i=1}^{N} \left\| \hat{\mathbf{y}}_i^t - \mathbf{y}_i^t \right\|^2}
\end{equation}
where $\hat{\mathbf{y}}$ and $\mathbf{y}$ are the predicted and ground truth field values at node $i$ and timestep $t$, and $\|\cdot\|$ denotes the Euclidean norm over $D$ variables.

\paragraph{Relative Root Mean Squared Error (RRMSE).} The RRMSE normalizes the RMSE by the root mean square of the ground truth, making it unitless and comparable across variables with different physical scales:
\begin{equation}
\mathrm{RRMSE}(t) = \frac{\mathrm{RMSE}(t)}{\mathrm{RMS}(\mathbf{y}^t)}, \qquad \mathrm{RMS}(\mathbf{y}^t) = \sqrt{\frac{1}{N} \sum_{i=1}^{N} \left\| \mathbf{y}_i^t \right\|^2}
\end{equation}
An RRMSE of 0 indicates perfect prediction; values less than 1.0 indicate the model error is smaller than the signal magnitude.

\paragraph{Normalized Mean Squared Error (NMSE).} The NMSE normalizes the mean squared error by the variance of the ground truth field, providing a scale-invariant measure of prediction quality.

\paragraph{Structural Similarity Index (SSIM).} SSIM measures perceptual similarity between predicted and ground truth fields, capturing luminance, contrast, and structural information simultaneously.

\paragraph{Nash--Sutcliffe Efficiency (NSE)}
The Nash--Sutcliffe Efficiency measures predictive skill by comparing the squared prediction error against the variance of the observed water depth field. It is defined as
\begin{equation}
    \text{NSE} = 1 - \frac{\displaystyle\sum_{i=1}^{N} 
    \left(\hat{h}_i - h_i\right)^2}
    {\displaystyle\sum_{i=1}^{N} \left(h_i - \bar{h}\right)^2}
    \label{eq:nse}
\end{equation}
where $\hat{h}_i$ and $h_i$ denote the predicted and reference water depths at node $i$, $N$ is the total number of nodes, and $\bar{h}$ is the mean reference water depth over all nodes. Higher NSE indicates better predictive performance, with $\text{NSE} = 1$ corresponding to perfect agreement, and values below zero indicating that the prediction is worse than using the mean observed depth as a predictor.

\paragraph{Critical Success Index (CSI)}
The Critical Success Index evaluates the model's ability to correctly identify flood exceedance events above a prescribed water-depth threshold. It is defined as
\begin{equation}
    \text{CSI} = \frac{H_{\text{thr}}}{H_{\text{thr}} + 
    M_{\text{thr}} + FA_{\text{thr}}}
    \label{eq:csi}
\end{equation}
where $H_{\text{thr}}$ is the number of hits, $M_{\text{thr}}$ is the number of misses, and $FA_{\text{thr}}$ is the number of false alarms for a threshold $h_{\text{thr}}$. Here, a hit corresponds to a node where both the prediction and the reference solution exceed $h_{\text{thr}}$, a miss corresponds to a node where only the reference solution exceeds the threshold, and a false alarm corresponds to a node where only the prediction exceeds the threshold. Higher CSI indicates better agreement in identifying significant flood events. Following the fluvial hydrology evaluation protocol, CSI can be reported at thresholds of 0.05 m and 0.30 m to capture both low-magnitude inundation and more operationally significant flooding.

\section{Per-Variable Model Predictions}

\subsection{Fluvial Hydrology}
\begin{table}[ht]
\centering
\begin{tabular}{l cc cc}
\toprule
& \multicolumn{2}{c}{\textbf{All Nodes}}
& \multicolumn{2}{c}{\textbf{Important Nodes}} \\
\cmidrule(lr){2-3} \cmidrule(lr){4-5}
Model
  & NSE $(\uparrow)$ & CSI $(\uparrow)$
  & NSE $(\uparrow)$ & CSI $(\uparrow)$ \\
\midrule
G-PARC           & \textbf{0.9402} & 0.8330          & \textbf{0.9176} & 0.8900 \\
G-PARC (\textbf{w/o MLS}) & 0.9171          & 0.8030          & 0.8683          & 0.8470 \\
MeshGraphKAN     & 0.9125          & \textbf{0.8630} & 0.8510          & \textbf{0.9030} \\
MeshGraphNet     & $-8543.10$      & 0.1980          & $-3937.10$      & 0.6030 \\
GraphSAGE        & $-971.77$       & $-0.228$        & $-688.35$       & 0.7690 \\
\bottomrule
\end{tabular}
\caption{\textbf{Aggregate fluvial hydrology metrics.} \textit{All Nodes} reports metrics over the full mesh; \textit{Important Nodes} restricts to nodes with depth $> 0.3$\,m at any timestep (median 27.8\% of nodes). Best result per column in bold.}
\label{tab:fluvial_comparison}
\end{table}

\begin{figure}[!h]
    \centering
    \includegraphics[width=\linewidth]{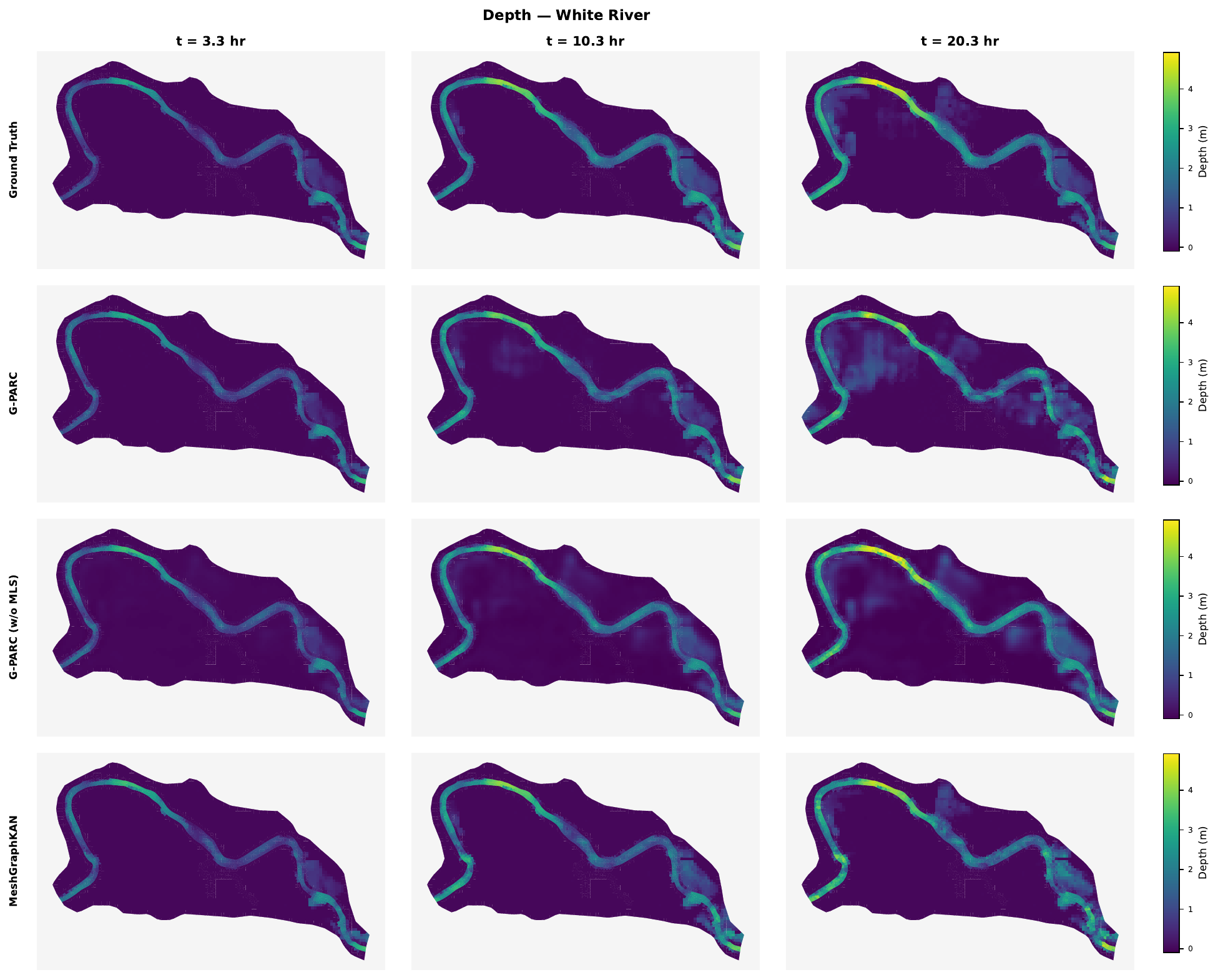}
    \caption{\textbf{Top three best performing model for White River Flooding Model Prediction: G-PARC, G-PARC (\textbf{w/o MLS}), MGKAN.}}
    \label{fig:WhiteRiver}
\end{figure}

\begin{landscape}
\begin{table}[p]
\centering
\small

\begin{tabular}{lccccc | >{\bfseries}c} 
\toprule
Metric & G-PARC & G-PARC (\textbf{w/o MLS}) & MGKAN & MGNET & GraphSAGE & Rel. Imp. \\
\midrule
RRMSE AUC $(\downarrow)$      & \textbf{0.2075 $\pm$ 0.1056} & 0.2995 $\pm$ 0.0964 & 0.2263 $\pm$ 0.1853 & 111.8 $\pm$ 30.01 & 18.61 $\pm$ 4.9516 & $-8.3\%$ \\
RRMSE$_{fin}$ $(\downarrow)$  & 0.3643 $\pm$ 0.2518          & 0.4620 $\pm$ 0.1705 & \textbf{0.3192 $\pm$ 0.2248} & 219.5 $\pm$ 72.67 & 32.86 $\pm$ 9.9446 & $-12.4\%$ \\
NMSE AUC $(\downarrow)$       & \textbf{0.0703 $\pm$ 0.0697} & 0.1196 $\pm$ 0.0770 & 0.1019 $\pm$ 0.1371 & 18065.8 $\pm$ 10125.9 & 469.6 $\pm$ 260.1 & $-31.0\%$ \\
SSIM AUC $(\uparrow)$         & \textbf{0.9521 $\pm$ 0.0403} & 0.9144 $\pm$ 0.0301 & 0.9474 $\pm$ 0.0615 & $5\times10^{-4} \pm 7\times10^{-4}$ & 0.0228 $\pm$ 0.0136 & $+0.0047$ \\
$R^2$ $(\uparrow)$            & \textbf{0.9093 $\pm$ 0.1007} & 0.8818 $\pm$ 0.0767 & 0.8856 $\pm$ 0.1399 & $-18586.9 \pm 9101.7$ & $-481.9 \pm 233.0$ & $+0.0237$ \\
RMSE AUC $(\downarrow)$       & \textbf{283.1 $\pm$ 236.7}  & 343.2 $\pm$ 220.4 & 327.9 $\pm$ 313.8 & 115256 $\pm$ 57160.8 & 18853.0 $\pm$ 8773.3 & $-13.6\%$ \\
\bottomrule
\end{tabular}
\caption[\textbf{Model performance comparison on the Fluvial Hydrology dataset.}]{\textbf{Comparison of model performance on the Fluvial Hydrology dataset ($n = 150$).} Best values per metric are bolded. Arrows ($\downarrow, \uparrow$) indicate direction of improvement. Rel. improvement compares G-PARC to the best performing baseline for each metric.}
\label{tab:river}

\vspace{1em} 

\begin{tabular}{llccccc | >{\bfseries}c}
\toprule
Metric & Variable & G-PARC & G-PARC (\textbf{w/o MLS}) & MGKAN & MGNET & GraphSAGE & Rel. Imp. \\
\midrule
\multirow{4}{*}{RRMSE AUC $(\downarrow)$} 
 & Depth      & \textbf{0.1749} & 0.2385 & 0.1841 & 58.63  & 23.83 & $-5.0\%$ \\
 & Volume     & \textbf{0.2075} & 0.2995 & 0.2263 & 111.8  & 18.61 & $-8.3\%$ \\
 & Velocity X & \textbf{0.2159} & 0.3142 & 0.2214 & 259.4  & 156.4 & $-2.5\%$ \\
 & Velocity Y & \textbf{0.2125} & 0.3366 & 0.2130 & 439.3  & 175.7 & $-0.2\%$ \\
\midrule
\multirow{4}{*}{RRMSE\textsubscript{fin} $(\downarrow)$} 
 & Depth      & 0.2931          & 0.3565 & \textbf{0.2591} & 123.6  & 43.34 & $-11.6\%$ \\
 & Volume     & 0.3643          & 0.4620 & \textbf{0.3192} & 219.5  & 32.86 & $-12.4\%$ \\
 & Velocity X & 0.3241          & 0.4301 & \textbf{0.2853} & 486.2  & 271.9 & $-12.0\%$ \\
 & Velocity Y & 0.2888          & 0.4847 & \textbf{0.2663} & 937.5  & 310.1 & $-7.8\%$ \\
\midrule
\multirow{4}{*}{NMSE AUC $(\downarrow)$} 
 & Depth      & \textbf{0.0562} & 0.0908 & 0.0763 & 7188.4  & 965.8  & $-26.3\%$ \\
 & Volume     & \textbf{0.0798} & 0.1331 & 0.1154 & 19757.8 & 513.2  & $-30.8\%$ \\
 & Velocity X & \textbf{0.0698} & 0.1343 & 0.0957 & 102475  & 33999.7 & $-27.1\%$ \\
 & Velocity Y & \textbf{0.0600} & 0.1453 & 0.0813 & 320992  & 40071.0 & $-26.2\%$ \\
\midrule
\multirow{4}{*}{SSIM AUC $(\uparrow)$} 
 & Depth      & \textbf{0.9609} & 0.9374 & 0.9557 & 0.0015          & 0.0505  & $+0.0052$ \\
 & Volume     & \textbf{0.9437} & 0.8870 & 0.9379 & 0.0011          & 0.0402  & $+0.0058$ \\
 & Velocity X & \textbf{0.9506} & 0.9206 & 0.9461 & $3\times10^{-4}$ & 0.0019  & $+0.0045$ \\
 & Velocity Y & \textbf{0.9530} & 0.9125 & 0.9499 & $-9\times10^{-4}$ & $-0.0015$ & $+0.0031$ \\
\midrule
\multirow{4}{*}{RMSE AUC $(\downarrow)$} 
 & Depth      & \textbf{0.1217} & 0.1455 & 0.1425 & 43.30  & 14.08   & $-14.6\%$ \\
 & Volume     & \textbf{566.2}  & 686.4  & 655.7  & 230513 & 37705.9 & $-13.6\%$ \\
 & Velocity X & \textbf{0.0277} & 0.0367 & 0.0291 & 33.33  & 19.49   & $-4.8\%$ \\
 & Velocity Y & \textbf{0.0252} & 0.0378 & 0.0274 & 57.54  & 21.00   & $-8.0\%$ \\
\bottomrule
\end{tabular}
\caption[\textbf{Per-variable model predictions on the Fluvial Hydrology dataset.}]{\textbf{Per-variable model predictions on the Fluvial Hydrology dataset.} Relative improvement shown per variable.}
\label{tab:river_pv}

\end{table}
\end{landscape}

\begin{figure}[!h]
    \centering
    \includegraphics[width=\linewidth]{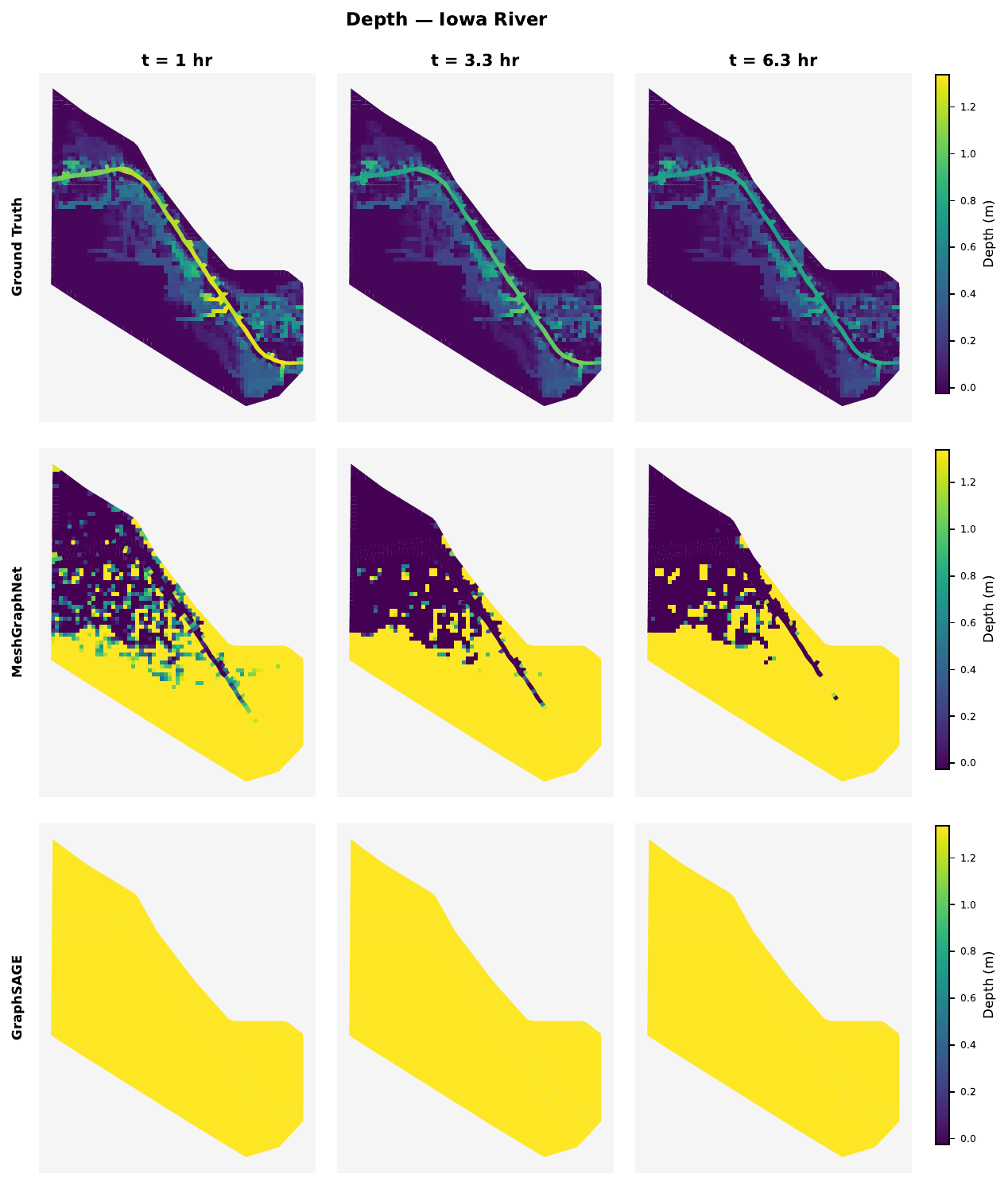}
    \caption{\textbf{Worst two baseline models for Iowa River Flooding Model Prediction: MGNET and GraphSAGE.}}
    \label{fig:riverexplosion}
\end{figure}
\FloatBarrier
\subsection{Planar Shock Wave}

\begin{landscape}
\begin{table}[p]
\centering
\small

\begin{tabular}{lccccc | >{\bfseries}c} 
\toprule
Metric & G-PARC & G-PARC (\textbf{w/o MLS}) & MGKAN & MGNET & GraphSAGE & Rel. Imp. \\
\midrule
RRMSE AUC $(\downarrow)$      & \textbf{0.0070 $\pm$ 0.0031} & 0.0824 $\pm$ 0.0662 & 0.0555 $\pm$ 0.0867 & 0.3544 $\pm$ 0.1708 & 7.5062 $\pm$ 3.3600 & $7.9\times$ \\
RRMSE\textsubscript{fin} $(\downarrow)$  & \textbf{0.0121 $\pm$ 0.0064} & 0.1222 $\pm$ 0.1021 & 0.1248 $\pm$ 0.1985 & 0.7470 $\pm$ 0.3550 & 13.60 $\pm$ 6.1487 & $10.1\times$ \\
NMSE AUC $(\downarrow)$       & \textbf{1.00$\times$10$^{-4} \pm$ 2.00$\times$10$^{-4}$} & 0.0265 $\pm$ 0.0470 & 0.0393 $\pm$ 0.0937 & 0.4448 $\pm$ 0.3634 & 178.7 $\pm$ 207.8 & $265.0\times$ \\
SSIM AUC $(\uparrow)$         & \textbf{0.9999 $\pm$ 2.00$\times$10$^{-4}$} & 0.9774 $\pm$ 0.0345 & 0.9881 $\pm$ 0.0244 & 0.8206 $\pm$ 0.1078 & 0.1585 $\pm$ 0.0447 & $+0.0118$ \\
$R^2$ $(\uparrow)$            & \textbf{0.9998 $\pm$ 2.00$\times$10$^{-4}$} & 0.9704 $\pm$ 0.0495 & 0.9588 $\pm$ 0.0964 & 0.4484 $\pm$ 0.4888 & $-198.6 \pm 204.3$ & $+0.0294$ \\
RMSE AUC $(\downarrow)$       & \textbf{0.0028 $\pm$ 8.00$\times$10$^{-4}$} & 0.0314 $\pm$ 0.0188 & 0.0281 $\pm$ 0.0515 & 0.1515 $\pm$ 0.0676 & 2.8872 $\pm$ 0.0711 & $10.0\times$ \\
\bottomrule
\end{tabular}
\caption{\textbf{Comparison of model performance on the planar shock wave dataset ($n = 75$).} Best values per metric are bolded. Arrows indicate direction of improvement.}
\label{tab:shock}

\vspace{3em} 

\begin{tabular}{llccccc | >{\bfseries}c}
\toprule
Metric & Variable & G-PARC & G-PARC (\textbf{w/o MLS}) & MGKAN & MGNET & GraphSAGE & Rel. Imp. \\
\midrule
\multirow{3}{*}{RRMSE AUC $(\downarrow)$} 
 & Density      & \textbf{0.0050} & 0.0545 & 0.0133 & 0.2199 & 12.91  & $2.7\times$ \\
 & $x$-Momentum & \textbf{0.0177} & 0.2034 & 0.1231 & 0.8642 & 5.7274 & $7.0\times$ \\
 & Total Energy & \textbf{0.0041} & 0.0539 & 0.0200 & 0.1889 & 6.8840 & $4.9\times$ \\
\midrule
\multirow{3}{*}{RRMSE\textsubscript{fin} $(\downarrow)$} 
 & Density      & \textbf{0.0091} & 0.0744 & 0.0277 & 0.4633 & 25.19  & $3.0\times$ \\
 & $x$-Momentum & \textbf{0.0202} & 0.1971 & 0.2249 & 1.4140 & 8.3667 & $9.8\times$ \\
 & Total Energy & \textbf{0.0076} & 0.0901 & 0.0432 & 0.3637 & 12.85  & $5.7\times$ \\
\midrule
\multirow{3}{*}{NMSE AUC $(\downarrow)$} 
 & Density      & \textbf{1$\times$10$^{-4}$} & 0.0118 & 0.0014 & 0.2384 & 733.9  & $14.0\times$ \\
 & $x$-Momentum & \textbf{6$\times$10$^{-4}$} & 0.1554 & 0.1246 & 2.0691 & 69.60  & $207.7\times$ \\
 & Total Energy & \textbf{1$\times$10$^{-4}$} & 0.0146 & 0.0034 & 0.1531 & 216.5  & $34.0\times$ \\
\midrule
\multirow{3}{*}{SSIM AUC $(\uparrow)$} 
 & Density      & \textbf{1.0000} & 0.9944 & 0.9993 & 0.9060 & 0.1668 & $+0.0007$ \\
 & $x$-Momentum & \textbf{0.9996} & 0.9445 & 0.9666 & 0.6382 & 0.0661 & $+0.0330$ \\
 & Total Energy & \textbf{1.0000} & 0.9932 & 0.9984 & 0.9177 & 0.2425 & $+0.0016$ \\
\midrule
\multirow{3}{*}{RMSE AUC $(\downarrow)$} 
 & Density      & \textbf{0.0017} & 0.0184 & 0.0042 & 0.0702 & 3.7954 & $2.5\times$ \\
 & $x$-Momentum & \textbf{0.0039} & 0.0434 & 0.0456 & 0.2287 & 1.4266 & $11.1\times$ \\
 & Total Energy & \textbf{0.0020} & 0.0250 & 0.0090 & 0.0955 & 2.9101 & $4.5\times$ \\
\bottomrule
\end{tabular}
\caption[\textbf{Per-variable metrics for the planar shock wave flow.}]{\textbf{Per-variable metrics for the planar shock wave flow.} Relative improvement shown per variable.}
\label{tab:shock_pv}

\end{table}
\end{landscape}

\begin{figure}[!h]
    \centering
    \includegraphics[width=\linewidth]{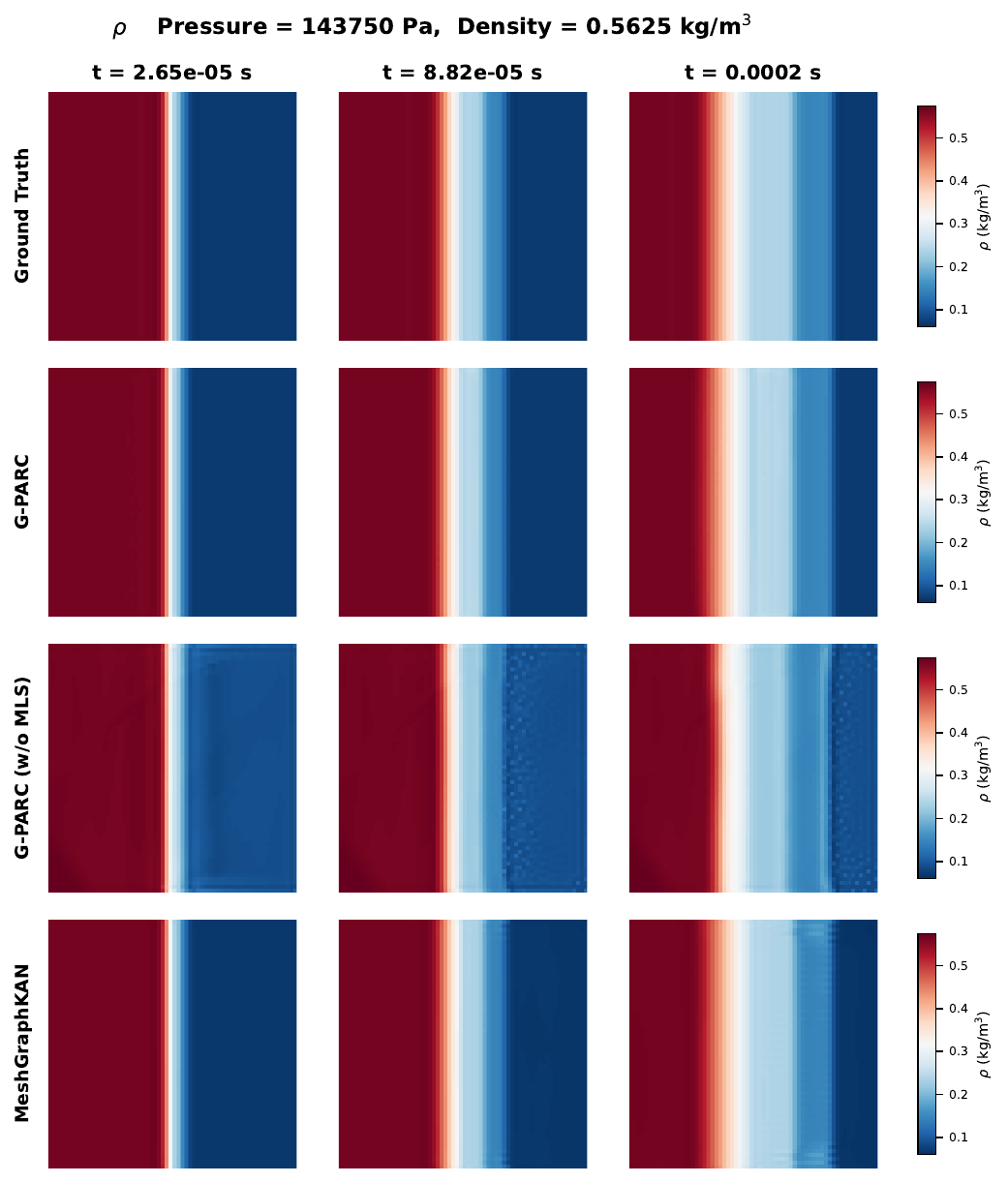}
    \caption{\textbf{Predicted density fields for the top three models (G-PARC, G-PARC (\textbf{w/o MLS}), and MGKAN) on the same representative test case as Figure~\ref{fig:shocktop}.}}
    \label{fig:density}
\end{figure}

\begin{figure}[!h]
    \centering
    \includegraphics[width=\linewidth]{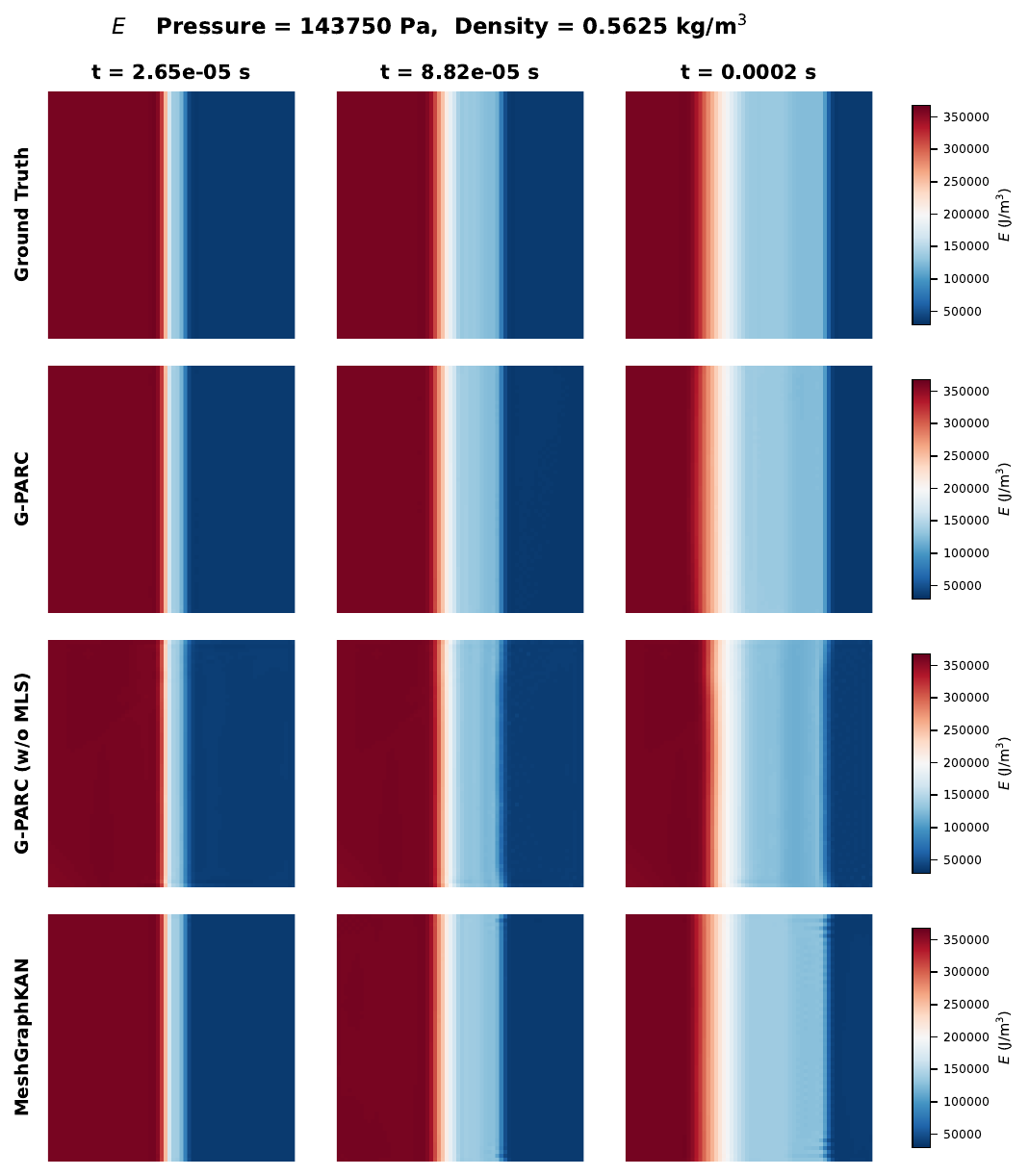}
    \caption{\textbf{Predicted total energy fields for the top three models (G-PARC, G-PARC (\textbf{w/o MLS}), and MGKAN) on the same representative test case as Figure~\ref{fig:shocktop}.}}
    \label{fig:energy}
\end{figure}

\begin{figure}[!h]
    \centering
    \includegraphics[width=\linewidth]{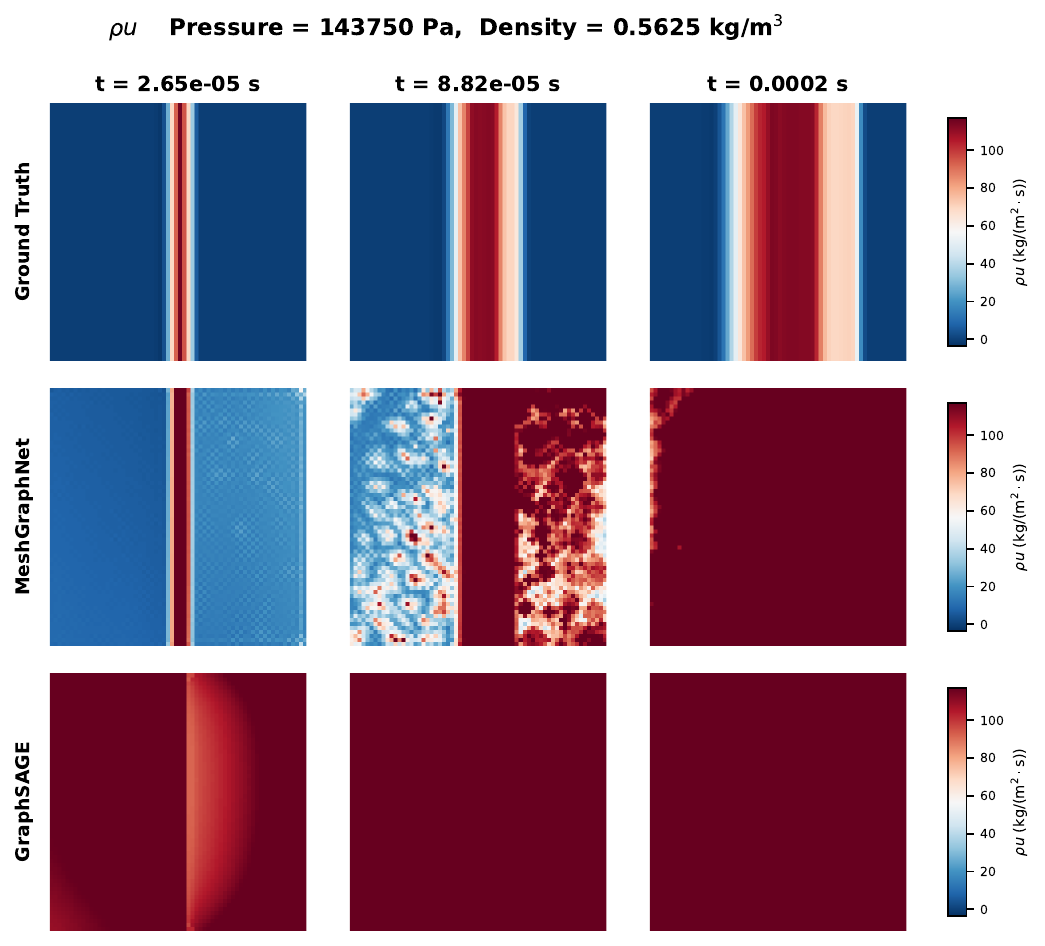}
    \caption{\textbf{Predicted $x$-momentum fields for the two worst-performing models (MGNET and GraphSAGE) on the same representative test case as Figure~\ref{fig:shocktop}, illustrating catastrophic rollout failure under long-horizon autoregressive prediction.}}
    \label{fig:worstxmoment}
\end{figure}
\FloatBarrier
\subsection{Elastoplastic}

\begin{figure}
    \centering
    \includegraphics[width=\linewidth]{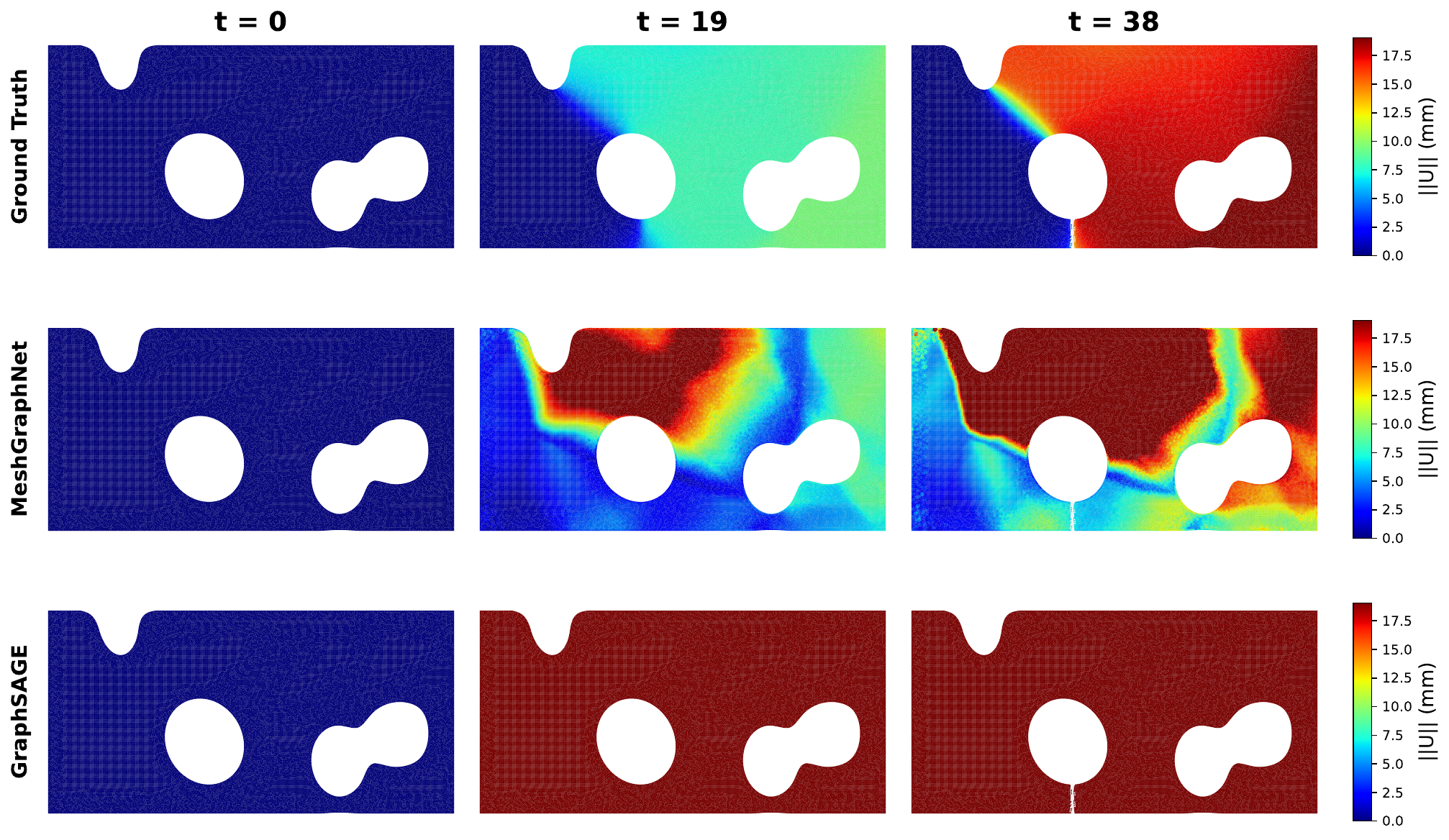}
    \caption{\textbf{Worst two performing models for elastoplastic dataset: MGNET and GraphSAGE.}}
    \label{fig:elastoexplo}
\end{figure}

\begin{landscape}
\begin{table}[p] 
\centering
\small

\begin{tabular}{lccccc | >{\bfseries}c} 
\toprule
Metric & G-PARC & G-PARC (\textbf{w/o MLS}) & MGKAN & MGNET & GraphSAGE & Rel. Imp. \\
\midrule
RRMSE AUC $(\downarrow)$      & \textbf{0.4391 $\pm$ 0.1328} & 0.4718 $\pm$ 0.1046 & 0.5850 $\pm$ 0.0981 & 3.4528 $\pm$ 1.0786 & 244.8 $\pm$ 71.09 & $-6.9\%$ \\
RRMSE\textsubscript{fin} $(\downarrow)$  & \textbf{0.4607 $\pm$ 0.1882} & 0.5240 $\pm$ 0.1300 & 0.5986 $\pm$ 0.1144 & 10.52 $\pm$ 4.0148 & 483.3 $\pm$ 145.8 & $-12.1\%$ \\
NMSE AUC $(\downarrow)$       & \textbf{0.2950 $\pm$ 0.1875} & 0.3435 $\pm$ 0.1682 & 0.5122 $\pm$ 0.2165 & 26.29 $\pm$ 16.42 & 127792 $\pm$ 66913.1 & $-14.1\%$ \\
SSIM AUC $(\uparrow)$         & \textbf{0.4346 $\pm$ 0.0899} & 0.4229 $\pm$ 0.1123 & 0.3336 $\pm$ 0.1204 & $-0.0798$ $\pm$ 0.0666 & 0.0392 $\pm$ 0.0129 & $+0.0117$ \\
$R^2$ $(\uparrow)$            & \textbf{0.6088 $\pm$ 0.2718} & 0.4958 $\pm$ 0.3149 & 0.3258 $\pm$ 0.3742 & $-66.30$ $\pm$ 37.27 & $-275395$ $\pm$ 116962 & $+0.1130$ \\
RMSE AUC $(\downarrow)$       & \textbf{1.9795 $\pm$ 0.7019} & 2.2724 $\pm$ 0.9127 & 2.7213 $\pm$ 1.0535 & 19.38 $\pm$ 2.6334 & 1440.5 $\pm$ 15.03 & $-12.9\%$ \\
\bottomrule
\end{tabular}
\caption{\textbf{Comparison of model performance on the Elastoplastic dataset ($n = 932$).} Best values per metric are bolded. Arrows indicate direction of improvement.}
\label{tab:elasto}

\vspace{3em} 

\begin{tabular}{llccccc | >{\bfseries}c}
\toprule
Metric & Var. & G-PARC & G-PARC (\textbf{w/o MLS}) & MGKAN & MGNET & GraphSAGE & Rel. Imp. \\
\midrule
\multirow{2}{*}{RRMSE AUC $(\downarrow)$} & $U_x$ & \textbf{0.4058} & 0.4392 & 0.5510 & 1.3841 & 135.8 & $-7.6\%$ \\
 & $U_y$ & \textbf{1.2581} & 1.4251 & 1.5117 & 40.47 & 2894.9 & $-11.7\%$ \\
\midrule
\multirow{2}{*}{RRMSE\textsubscript{fin} $(\downarrow)$} & $U_x$ & \textbf{0.4384} & 0.5063 & 0.5835 & 1.6365 & 241.3 & $-13.4\%$ \\
 & $U_y$ & 1.3591 & \textbf{1.2577} & 1.2616 & 143.1 & 5882.4 & $-0.3\%$ \\
\midrule
\multirow{2}{*}{NMSE AUC $(\downarrow)$} & $U_x$ & \textbf{0.5115} & 0.6183 & 0.9584 & 4.8777 & 58792.9 & $-17.3\%$ \\
 & $U_y$ & \textbf{3.4076} & 5.7001 & 5.1047 & 9841.3 & 49244283 & $-33.2\%$ \\
\midrule
\multirow{2}{*}{SSIM AUC $(\uparrow)$} & $U_x$ & \textbf{0.7903} & 0.7481 & 0.6209 & $-0.1735$ & 0.0649 & $+0.0422$ \\
 & $U_y$ & \textbf{0.0789} & 0.0978 & 0.0463 & 0.0139 & 0.0136 & $+0.0189$ \\
\midrule
\multirow{2}{*}{RMSE AUC $(\downarrow)$} & $U_x$ & \textbf{2.5978} & 3.0334 & 3.6839 & 8.7687 & 1081.1 & $-14.4\%$ \\
 & $U_y$ & \textbf{0.8689} & 0.8886 & 0.9199 & 25.18 & 1720.3 & $-2.2\%$ \\
\bottomrule
\end{tabular}
\caption{\textbf{Per-component metrics for the Elastoplastic dataset.} Relative improvement shown per displacement component.}
\label{tab:elasto_pv}

\end{table}
\end{landscape}

\end{document}